\documentclass[10pt]{article} 
\usepackage{pifont}
\usepackage{wrapfig}
\usepackage{caption}
\newcommand{\cmark}{\ding{51}}
\newcommand{\xmark}{\ding{55}}
\newcommand{\hlcell}[1]{\cellcolor{lightorchid}#1}
\usepackage{booktabs}
\usepackage{graphicx}
\usepackage{float}
\usepackage{siunitx}
\usepackage{float}
\usepackage[preprint]{tmlr}

\usepackage{amsmath,amsfonts,bm}

\def\eqref#1{equation~\ref{#1}}

\def\1{\bm{1}}

\DeclareMathAlphabet{\mathsfit}{\encodingdefault}{\sfdefault}{m}{sl}
\SetMathAlphabet{\mathsfit}{bold}{\encodingdefault}{\sfdefault}{bx}{n}

\usepackage{hyperref}
\usepackage{url}
\usepackage{algorithm}
\usepackage[accsupp]{axessibility} 
\usepackage{algorithmic}
\usepackage{makecell}
\usepackage{bbm}
\usepackage{pifont}
\usepackage{booktabs}
\usepackage{graphicx} 
\usepackage{multirow}
\usepackage{xcolor}
\usepackage[table]{xcolor}
\usepackage{colortbl}
\usepackage{xspace}
\usepackage{booktabs}
\usepackage{tabularx}
\usepackage{hyperref}
\usepackage[dvipsnames]{xcolor}

\newcommand{\std}[2]{#1$_{\scriptsize\pm#2}$}

\title{Single-Exam Mammography Risk Prediction with Privileged History Distillation}

\author{\name Banafsheh Karimian  \email banafsheh.karimian.1@ens.etsmtl.ca\\
      \addr LIVIA,  Dept. of Systems Engineering, ETS Montreal, Canada \\
      \AND
      \name Soufiane Belharbi   \email soufiane.belharbi@ens.etsmtl.ca\\
      \addr LIVIA,  Dept. of Systems Engineering, ETS Montreal, Canada
      \AND
      \name Alexis Guichemerre   \email alexis.guichemerre.1@ens.etsmtl.ca\\
      \addr LIVIA,  Dept. of Systems Engineering, ETS Montreal, Canada
      \AND
      \name Luke McCaffrey   \email luke.mccaffrey@mcgill.ca\\
      \addr Goodman Cancer Institute, Dept. of Oncology, McGill University, Canada
      \AND
      \name Mohammadhadi Shateri   \email mohammadhadi.shateri@ens.etsmtl.ca\\
      \addr LIVIA,  Dept. of Systems Engineering, ETS Montreal, Canada
      \AND
      \name Eric Granger   \email eric.granger@ens.etsmtl.ca\\
      \addr LIVIA,  Dept. of Systems Engineering, ETS Montreal, Canada}

\def\month{MM}  
\def\year{YYYY} 
\def\openreview{\url{https://openreview.net/forum?id=XXXX}} 

\begin{document}

\maketitle

\begin{abstract}
Longitudinal mammography screening has become an important source of information for improving future breast cancer risk prediction. However, the performance of current longitudinal mammography models degrades when prior examinations are unavailable at inference, creating a structured privileged-information setting in which temporal context is available during training but absent at deployment. 
We propose Single-Exam Mammography risk prediction with privileged History Distillation (\ours), a framework that uses longitudinal history as privileged information available only during training to preserve the predictive benefits of longitudinal modeling while requiring only the current screening examination at deployment. During training, the student relies on the current examination to predict latent representations of prior visits, while horizon-specific teachers provide additional supervision from the observed longitudinal history. Together, latent history prediction and teacher distillation preserve the temporal modeling structure of longitudinal predictors under current-exam-only inference.
We validate \ours on three longitudinal mammography cohorts, the CSAW-CC, EMBED, and OMI-DB, using the transformer-based Longitudinal Mammography Risk (LoMaR) and recurrent Visual Memory Recurrent Attention (VMRA) backbones. Under current-exam-only inference, \ours consistently improves long-horizon AUC and pAUC over longitudinal models evaluated without history, particularly in the clinically relevant low false-positive-rate region. It also recovers much of the performance gap with respect to full-history inference across datasets and backbones. Ablations further show that these gains are not reproduced by masking or heuristic history imputation. The strongest performance is achieved by combining patient-specific latent history prediction with distilled temporal risk supervision. These results suggest that temporal context can be effectively exploited as privileged supervision while preserving the underlying longitudinal modeling structure. Code available: \href{https://github.com/BanafshehKarimian/PHD}{link.}
\end{abstract}    
\section{Introduction}
\label{sec:intro}

Breast cancer is one of the most common cancers worldwide and remains a major cause of cancer-related mortality \cite{Hakama2008-fi,Wilkinson2022-dg,Bray2024-sv,Grasso2025}. Detecting breast cancer at earlier stages through screening is strongly associated with improved prognosis and reduced mortality \cite{Ginsburg2020}. Mammography is one of the primary imaging modalities for breast cancer screening due to its relatively low cost, wide accessibility, and established clinical utility \cite{Smith2003,LaubySecretan2015,Wilkinson2025-nx}. Beyond detecting existing malignancies, estimating a woman's future risk of developing breast cancer has become an increasingly important component of personalized screening programs, enabling risk-adapted screening intervals, earlier intervention, and more efficient allocation of healthcare resources. Recent deep learning methods have substantially improved prediction of future breast cancer risk directly from screening mammograms, outperforming traditional statistical risk models based on demographic and clinical variables \cite{Kim2021-cy,Kim2023-ig,Santeramo2024-zx,Mendes2025-br,thrun2025reconsidering}. A representative example is Mirai~\cite{Yala2021}, which predicts multi-year risk from a single screening examination.

However, such single-exam methods are limited because they do not account for the temporal aspect of disease development. Indeed, cancer risk prediction is inherently a temporal problem: previous screening mammograms carry clinically important information, and changes across successive exams (appearance, disappearance of subtle findings, additional evidence of disease development). Consequently, using longitudinal screening history has become an important direction for improving risk prediction \cite{lomar, vmra}. 
More recent approaches, including LoMaR~\cite{lomar} and VMRA \cite{vmra}, model longitudinal screening trajectories by aggregating information across multiple exams. These methods consistently show improved long-horizon risk prediction by exploiting temporal information unavailable in single-exam models, especially for longer-horizon risk prediction.

\begin{wrapfigure}{r}{0.53\columnwidth}
    \centering
    \vspace{-8pt}
    \includegraphics[width=\linewidth]{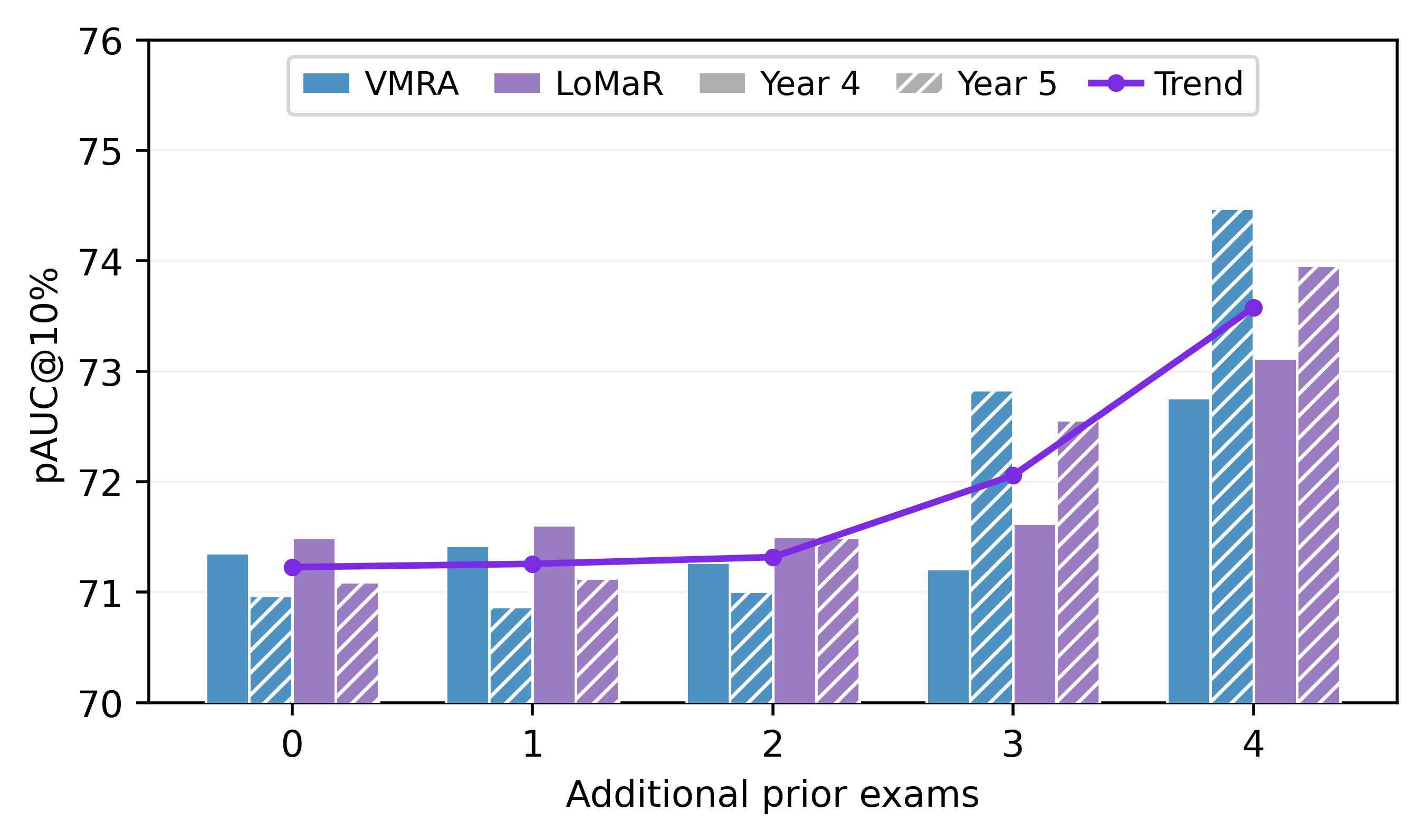}
    \captionsetup{font=small}
    \caption{pAUC@10\% at the 4- and 5-year horizons for LoMaR and VMRA trained with full longitudinal history on CSAW-CC and evaluated with only 0 to 4 prior exams used alongside the current exam, if available. Performance declines as prior exams are masked, indicating the reliance of longitudinally trained models on screening history at inference time.}
    \label{fig:history}
\end{wrapfigure}

Despite these advances, existing longitudinal approaches rely on an assumption that limits their practical deployment: they require previous screening examinations both during training and at inference time~\cite{lomar, vmra}. In routine clinical practice, however, historical mammograms are often incomplete or unavailable for several reasons: patients change healthcare providers, undergo their first screening examination, miss screening rounds, or have images stored across different institutions \cite{Reece2021}. Prior work has shown that removing exam history at inference substantially reduces future risk prediction performance \cite{lomar}. As shown in Fig.~\ref{fig:history}, Longitudinal Mammography Risk (LoMaR) ~\cite{lomar} and Visual Memory Recurrent Attention (VMRA) ~\cite{vmra} perform better when additional prior screening exams are available. Masking these exams at inference reduces predictive accuracy, particularly at longer prediction horizons. This highlights the importance of screening history at inference, as longitudinally trained models perform better when prior examinations are available.

This discrepancy raises a fundamental question: \textbf{Can a partially observed temporal history available during training be used to improve a model that receives only the current observation at inference?} This problem is particularly relevant in longitudinal mammography, where available prior screening examinations contain useful temporal information but may be unavailable at deployment. Rather than treating screening history as information that must be available at inference for the model to perform well, we instead use it as privileged supervision during training to improve a model that ultimately operates from the current examination alone. Such a method would exploit the predictive advantages of longitudinal supervision while retaining the practicality of single-exam inference. This perspective naturally connects the problem to the Learning Using Privileged Information (LUPI) paradigm \cite{vapnik2015}, where privileged information is accessible during training but unavailable at deployment.

Motivated by this observation, we propose \textbf{Single-Exam Mammography risk prediction with privileged History Distillation (\ours)}, a method that treats historical mammograms as privileged information. During training, horizon-specific teacher models, which have access to the full available history, learn temporal risk representations. In contrast, the student model receives only the current screening examination. Instead of attempting to reconstruct missing mammograms, since absent image information cannot be perfectly recovered, the student predicts compact latent representations containing from the current examination. Feature-level supervision and horizon-specific distillation then transfer temporal knowledge from the longitudinal teachers into the single-exam student. Consequently, the final model retains the deployment simplicity of single-exam inference while benefiting from additional information provided by temporal information available only during training. 

The main contributions of this work are threefold. First, we formulate a structured longitudinal LUPI setting in which an irregular, partially observed temporal sequence is available during training but deployment is restricted to the current observation, and instantiate this setting for mammography risk prediction. Second, we introduce \ours, which preserves the temporal modeling structure of an existing longitudinal predictor by estimating unavailable time-indexed latent representations from the current observation and transferring horizon-specific risk information from privileged longitudinal teachers. Third, we show that this formulation is compatible with distinct temporal aggregation mechanisms by integrating \ours with the transformer-based LoMaR and recurrent VMRA backbones across CSAW-CC, EMBED, and OMI-DB. Results show consistent improvements in long-horizon breast cancer risk prediction under missing-history inference, recovering much of the performance gap to full-history inference, particularly at longer horizons. 
\section{Related Work}
\label{sec:rw}

The literature related to our proposed approach can be broadly categorized into three directions: (i) future breast cancer risk prediction from mammography, (ii) longitudinal mammography risk prediction using multiple screening examinations, and (iii) learning with privileged information and knowledge distillation.
\subsection{Future Breast Cancer Risk Prediction:} Breast cancer risk prediction has traditionally relied on statistical models such as Tyrer-Cuzick \cite{himes2016breast} and the Breast Cancer Surveillance Consortium model \cite{tice2019validation}, which estimate risk from predefined demographic and clinical variables. Recent deep learning approaches learn risk directly from screening mammograms and have shown improved prediction \cite{Yala2021,Kim2023-ig,Mendes2025-br,Santeramo2024-zx,thrun2025reconsidering}. Among them, Mirai~\cite{Yala2021} introduced a multi-horizon deep learning formulation for predicting 1--5 year risk from a single screening examination. While effective for single-exam prediction, it does not model temporal changes across successive screenings, motivating longitudinal approaches that incorporate prior examinations.

\subsection{Longitudinal Mammography Risk Prediction:}
Recent work extended single-exam risk prediction to longitudinal screening histories. These methods differ mainly in how temporal information is represented and aggregated, grouped into spatially resolved multi-view approaches and visit-level temporal aggregation methods. Improvements of these methods are dependent to an important deployment assumption: prior screening exams are provided to the model at inference time.

\textbf{Spatially resolved multi-view modeling:}
Some longitudinal approaches preserve individual mammographic views and explicitly model temporal changes between corresponding prior and current images. LMV-Net~\cite{lmv}, for example, aligns prior and current CC and MLO feature maps using image-derived deformation fields and combines view-specific temporal representations through self- and cross-attention. Such explicit spatial comparison can preserve localized temporal changes, but it inherently requires corresponding prior mammograms at inference to construct the temporal alignments.

\textbf{Visit-level temporal aggregation:}
A second family first aggregates the four mammographic views of each screening examination into a compact visit-level representation and subsequently models the sequence of visits. LoMaR~\cite{lomar} extends the Mirai~\cite{Yala2021} formulation by processing these visit embeddings with a temporal transformer, demonstrating that additional screening history is particularly beneficial for longer-horizon risk prediction. MTP-BCR~\cite{wang2025mtp} similarly employs transformer-based longitudinal modeling while incorporating clinical and prognostic factors. VMRA~\cite{vmra} instead models the visit sequence using a Vision-Mamba~\cite{vmamba24} recurrent state-space architecture and incorporates asymmetry-aware representations to capture persistent bilateral differences over time.

Although these approaches differ in how they model temporal information, their longitudinal advantage depends on observed prior examinations being available at inference. When these examinations are missing, the temporal sequence or pairwise comparisons on which these models operate are incomplete or unavailable, creating a deployment mismatch between longitudinal training and current-exam-only inference. \ours addresses a different setting: rather than requiring longitudinal information as an inference-time input, it treats screening history as privileged training information.

\subsection{Learning with Privileged Information and Knowledge Distillation:} 

In the Learning Using Privileged Information (LUPI) paradigm \cite{vapnik2009,vapnik2015}, auxiliary information is available during training but unavailable at deployment. Knowledge distillation provides a practical mechanism for transferring such information from a privileged teacher to a restricted-input student \cite{hinton2015,lopezpaz2016unifying}. Privileged supervision has been explored across multimodal settings \cite{aslam2024distilling,aslam2024multi,mha}, while temporal information available only during training has also been used to improve models operating from restricted observations at inference \cite{karlsson2021using,davydov2024using}. More closely related temporal distillation methods transfer information from multi-frame or multi-scan teachers to single-frame models \cite{wang2020,qiu2023}, while missing-modality approaches estimate representations available during training but absent at deployment \cite{hong2025mutud}. 

The setting considered here differs in that the privileged information is a patient-specific, partially observed temporal history: different patients provide supervision for different prior time points, while deployment may contain no prior examinations. These unavailable observations jointly form the temporal input structure on which the longitudinal predictor operates. Direct distillation from a full-history teacher to a current-exam student does not explicitly preserve the longitudinal model's temporal input structure. In LUPI, such architectural differences can complicate the attribution of gains to privileged information \cite{provodin2025rethinking}. \ours therefore preserves the longitudinal modeling structure, predicts time-indexed representations of unavailable prior visits, and transfers horizon-specific risk information while requiring only the current examination at inference.

\section{Proposed Methodology}
\label{sec:poroposal}
\subsection{Problem Formulation and Notation}

We consider a supervised prediction setting in which a current observation is available at both training and inference, while a sequence of historical observations is available only during training. This defines a structured form of privileged information in which the unavailable input is an ordered temporal context rather than a single auxiliary feature or modality, where not all time slots have available data. We instantiate this setting for longitudinal mammography risk prediction as follows. Let $\Mammo{v}{\timeidx}$ denote mammogram image, $I$, with mammographic view $v$ acquired at relative screening time $\timeidx$. We define the set of standard mammographic views as 
$\ViewSet
=
\{
\text{L-CC},
\text{R-CC},
\text{L-MLO},
\text{R-MLO}
\},$
where L and R denote the left and right breasts, respectively, and CC and MLO denote the craniocaudal and mediolateral oblique views.

The complete set of relative screening-time indices is 
$\TimeSet
=
\{-\maxprior,\ldots,-1,0\},$
where $0$ denotes the current screening examination and negative indices denote prior screening visits, from $1$ to $\maxprior$ years ago. Since longitudinal histories may be incomplete and irregular, we let $\ObservedTimes\subseteq\TimeSet$ denote the set of observed screening times for each patient. The corresponding available screening history is
$\ScreenHistory
=
\left\{
\Mammo{v}{\timeidx}
\;:\;
\timeidx\in\ObservedTimes,\;
v\in\ViewSet
\right\}.
$
During training, the number of available historical examinations may vary across patients. Thus, only available visits are provided to the longitudinal pathways, and unavailable visits do not contribute to the feature-level objective, following \cite{lomar, vmra}.

For each future prediction horizon $\horidx$ (cancer by year $\horidx$), the target conditional risk is
\begin{equation}
\RiskProbability{\horidx}
=
\mathbb{P}
\left(
t_{\mathrm{cancer}}
\leq
\horidx
\mid
\ScreenHistory
\right),\qquad
\horidx\in\{1,\ldots,\numhorizons\},
\label{eq:problem-formulation}
\end{equation}
where $t_{\mathrm{cancer}}$ denotes the time until breast cancer diagnosis and $\numhorizons$ is the total number of future prediction years. The objective is to learn model predictions $\PredictedRisk{\horidx}$ that approximate $\RiskProbability{\horidx}$ for all prediction horizons.

State-of-the-art multi-visit risk prediction models estimate these probabilities by aggregating the current examination with prior visits. At inference time, however, prior mammograms may be unavailable. The missing-history inference setting considered in this work therefore corresponds to
$
\ObservedTimes^{\mathrm{test}}=\{0\}, 
$
meaning the model receives only the four views of the current examination at test time, or the possible number of available prior at test time, $\eta$, is $0$.

Our objective therefore is to exploit longitudinal screening information during training while predicting multi-horizon risk from the current examination alone at deployment. This setting follows the LUPI paradigm~\cite{vapnik2009}, in which privileged information is available during training but unavailable at inference. In our formulation, the privileged information consists of prior mammographic screening visits. \ours transfers temporal risk information learned from these longitudinal trajectories into a practical single-examination inference model.

\begin{figure*}[!t]
    \centering
    \includegraphics[width=\textwidth]{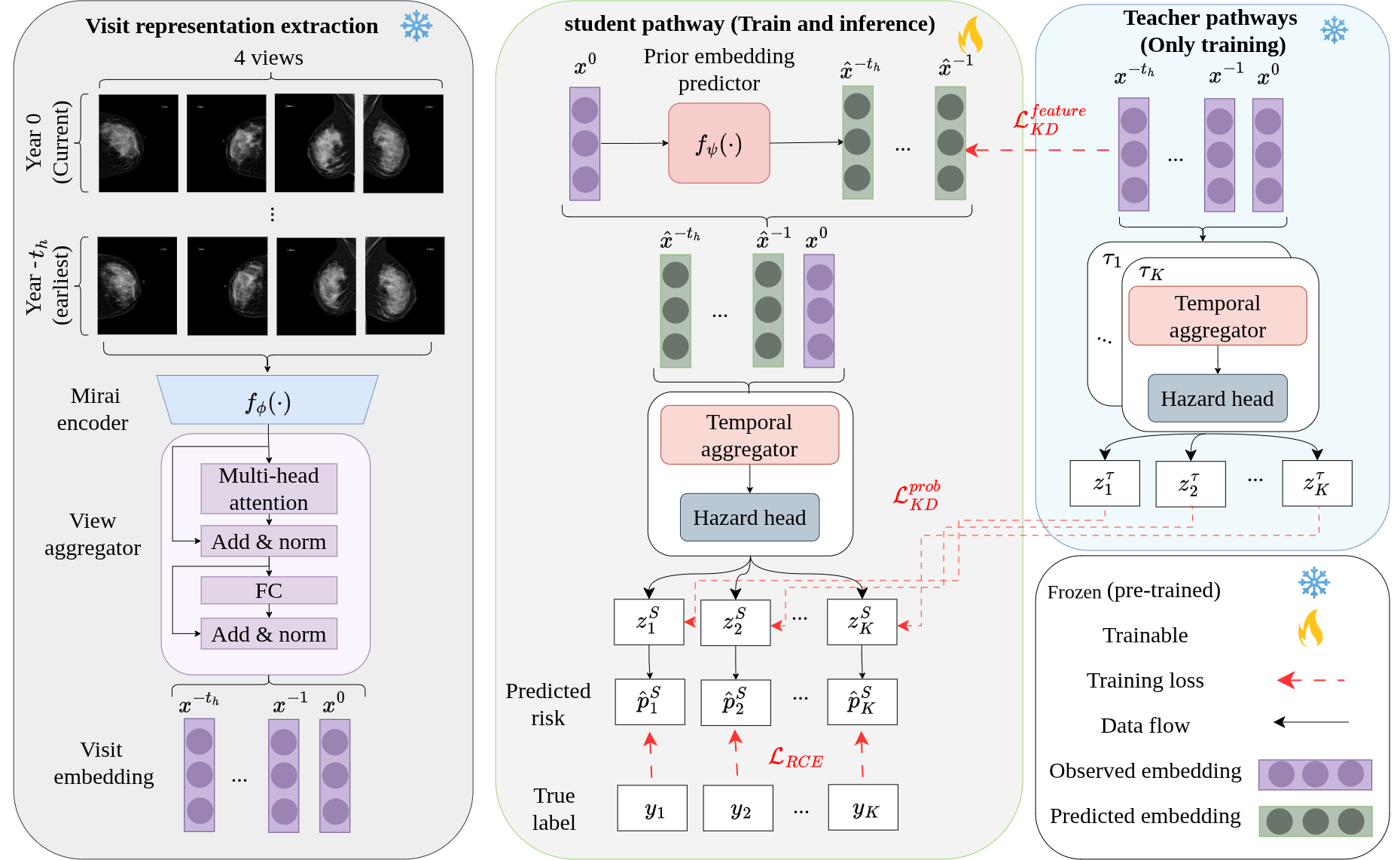}
    \caption{Overview of the proposed \ours framework. First representations are extracted. Then the student uses only the current screening exam, predicts latent embeddings for missing prior visits, and estimates multi-horizon risk. During training, horizon-specific teachers use the observed longitudinal history, while feature- and probability-level distillation transfers temporal information to the student. At inference, the teacher pathways are discarded, and no history is used.
}
    \label{fig:diagram}
\end{figure*}

\subsection{\ours Overview}

\ours follows a simple design principle: rather than replacing a longitudinal predictor with a structurally different single-observation model when history is unavailable, it preserves the longitudinal modeling interface and supplies it with predicted latent representations of the missing temporal context. The Fig.~\ref{fig:diagram} shows overall \ours. It consists of three main components: visit representation extraction, a single-exam student, and a set of privileged horizon-specific teachers. Following existing longitudinal mammography risk prediction methods \citep{lomar,vmra}, each screening examination is encoded into a compact visit-level representation using the frozen Mirai mammography encoder and view aggregation module. 

\ours considers the prior screening visits as privileged information: available during training but not at deployment. During training, the prior embedding predictor $\HistoryPredictor$ is optimized to estimate prior visit embeddings from the current visit embedding $\VisitEmbedding{0}$, using the corresponding observed prior embeddings as supervision whenever they are available. At deployment, the student receives only $\VisitEmbedding{0}$ and uses the trained predictor to generate $\PredVisitEmbedding{-1},\ldots,\PredVisitEmbedding{-\maxprior}$ as latent substitutes for the unavailable prior visits. The current and predicted prior embeddings are then processed by the temporal aggregation backbone $\TemporalAggregator$ and the additive hazard head to produce multi-horizon risk estimates.

During training, \ours uses independently trained horizon-specific teachers $\Teacher{1},\ldots,\Teacher{\numhorizons}$, each operating on the observed longitudinal sequence. Although a single teacher could jointly predict all horizons, shared multi-horizon optimization may introduce interference because the predictive evidence and class imbalance differ across short- and long-term horizons. Training each teacher for one horizon allows it to specialize in the corresponding temporal risk signal before its parameters are frozen for student distillation. Feature-level distillation aligns predicted and observed prior embeddings, while probability-level distillation transfers the specialized risk information from each teacher to its corresponding student output. After training, all teachers are discarded, and inference uses only the current examination as the input to the student.

\subsection{Visit Representation Extraction}

To obtain a compact representation for each screening examination, we use the pretrained Mirai mammography encoder \citep{Yala2021}, denoted by $\ImageEncoder(\cdot)$. For a screening visit at relative time $\timeidx$, each mammographic view $\Mammo{v}{\timeidx}$, with $v\in\ViewSet$, is processed independently
to produce a view-level representation:
$
\ViewFeature{v}{\timeidx}
=
\ImageEncoder\!\left(
    \Mammo{v}{\timeidx}
\right).
$
Following Mirai \cite{Yala2021}, the representations of the standard bilateral CC and MLO views from the same examination are combined using the pretrained $\ViewAggregator$:
$
\VisitEmbedding{\timeidx}
=
\ViewAggregator\!\left(
    \left\{
        \ViewFeature{v}{\timeidx}
    \right\}_{v\in\ViewSet}
\right),$
where $\VisitEmbedding{\timeidx}$ denotes the visit-level embedding at relative screening time $\timeidx$. Both the image encoder and the view aggregation module remain frozen throughout training, following \cite{lomar}. Applying this procedure to all observed screening visits yields the
patient-specific longitudinal embedding sequence
$
\left\{
    \VisitEmbedding{\timeidx}
\right\}_{\timeidx\in\ObservedTimes},
$
where $\ObservedTimes\subseteq\TimeSet$ denotes the set of screening times available for that patient. For a fully observed history, this sequence is
$
\left\{
    \VisitEmbedding{-\maxprior},
    \ldots,
    \VisitEmbedding{-1},
    \VisitEmbedding{0}
\right\}.
$

\ours does not require every patient to have a complete longitudinal screening history at training. Consistent with prior longitudinal mammography risk prediction frameworks \citep{lomar,vmra}, unavailable visits are masked during training. Consequently, only embeddings corresponding to $\timeidx\in\ObservedTimes$ are provided to the teacher pathways and included in the feature-level distillation objective. This formulation naturally supports variable-length and partially observed screening histories.

\subsection{Prior Embedding Prediction and Feature Distillation}

Given the current visit embedding $\VisitEmbedding{0}$, \ours trains a lightweight prior embedding predictor $\HistoryPredictor$ to estimate latent representations associated with the unavailable screening history:
$
\left\{
    \PredVisitEmbedding{\timeidx}
\right\}_{\timeidx=-1}^{-\maxprior}
=
\HistoryPredictor\!\left(
    \VisitEmbedding{0}
\right),
$
where $\PredVisitEmbedding{\timeidx}$ denotes the predicted visit-level embedding at relative screening time $\timeidx$. The prior embedding predictor is implemented as a lightweight multi-layer perceptron operating on the current visit embedding, predicting one output head per history year. It predicts the embedding associated with each prior visit directly from $\VisitEmbedding{0}$ rather than generating the historical sequence autoregressively, so that prediction errors at one historical position do not propagate to the remaining positions. \ours performs prediction entirely in latent representation space and does not attempt to reconstruct the corresponding mammographic images, avoiding high-resolution image generation. The additional inference-time computation is limited to the lightweight prior-embedding predictor.

During training, the predicted embeddings are supervised using the corresponding observed prior visit embeddings. Let $\ObservedHistoryTimes = \ObservedTimes \setminus \{0\}$ denote the patient-specific set of observed prior screening times. The feature-level distillation objective is
defined as:
\begin{equation}
\mathcal{L}_{\mathrm{KD}}^{\mathrm{feature}}
=
\frac{1}{\left|\ObservedHistoryTimes\right|}
\sum_{\timeidx\in\ObservedHistoryTimes}
\left\|
    \VisitEmbedding{\timeidx}
    -
    \PredVisitEmbedding{\timeidx}
\right\|_2^2.
\label{eq:feature-distillation}
\end{equation}
This loss is evaluated only for prior visits that are available for a given patient. Therefore, incomplete longitudinal histories are handled naturally without requiring a fully observed screening sequence. For patients with no observed prior examinations, the feature-distillation term is omitted.

\subsection{Longitudinal Aggregation and Risk Prediction}

\ours is agnostic to the temporal aggregation architecture and can be integrated with any longitudinal backbone that maps a sequence of visit embeddings to a unified temporal representation. In this work, we evaluate \ours with the transformer-based visit aggregator, LoMaR \citep{lomar} and the recurrent VMRNN encoder, VMRA \citep{vmra}. In both cases, \ours retains the original temporal architecture and introduces the prior embedding prediction and distillation mechanisms around it.

In the primary missing-history setting, the student receives the current visit embedding together with the predicted prior embeddings. We denote this sequence by
$
\StudentSequence
=
\left\{
    \VisitEmbedding{0},
    \PredVisitEmbedding{-1},
    \ldots,
    \PredVisitEmbedding{-\maxprior}
\right\}.
$
The temporal aggregation module then maps this sequence to a unified
student representation:
$
\TemporalRepresentation
=
\TemporalAggregator\!\left(
    \StudentSequence
\right),
$
where $\TemporalAggregator$ denotes either the transformer aggregator (LoMaR) or the recurrent one (VMRA). Following prior mammography risk prediction methods, risk is estimated using the standard additive hazard formulation \cite{lomar}. For each prediction horizon $\horidx\in\{1,\ldots,\numhorizons\}$, the model predicts a non-negative hazard increment $
l_{\horidx}^{\student}
=
\rho
\left(
    \mathbf{w}_{\horidx}^{\top}
    \TemporalRepresentation
    +
    b_{\horidx}
\right),
$
where $\rho(\cdot)$ is a non-negative activation function and $\mathbf{w}_{\horidx}$ and $b_{\horidx}$ are learnable horizon-specific parameters. The output logits $z^S_\horidx$ are defined as:
\begin{equation}
z^S_\horidx
=
B\!\left(\TemporalRepresentation\right)
+
\sum_{i=1}^{\horidx}
l_i^{\student}, 
\label{eq:cumulative-risk}
\end{equation}
where $B(\TemporalRepresentation)$ is the baseline term. Cumulative predicted risk up to horizon $\horidx$, is after defined as $\PredictedRisk{\horidx} = \sigma(z_\horidx)$, where $\sigma(\cdot)$ is the sigmoid function. To supervise multi-horizon risk prediction, let $\LabelVector\in\{0,1,-1\}^{\numhorizons}$ denote the label vector, where $\RiskLabel{\horidx}=-1$ indicates that the outcome at horizon $\horidx$ is unavailable, 0 indicates not malignant and 1 indicates malignant. Such labels are excluded through the mask
${
\LabelMask{\horidx}
=
\mathbb{1}
\left[
    \RiskLabel{\horidx}\neq -1
\right].
}$ Since cancer prevalence varies across prediction horizons, the Re-weighted Cross Entropy loss \citep{Karaman2022} is used. For each horizon, the positive-class weight is defined as
$
\ClassWeight{\horidx}
=
\frac{N_{\horidx}^{-}}{N_{\horidx}^{+}},
$
where $N_{\horidx}^{+}$ and $N_{\horidx}^{-}$ are the numbers of positive and negative training samples with valid labels at horizon $\horidx$, respectively. The supervised student loss is defined as:
\begin{equation}
\RCELoss
=
\frac{
    1
}{
    \sum_{\horidx=1}^{\numhorizons}
    \LabelMask{\horidx}
}
\sum_{\horidx=1}^{\numhorizons}
\LabelMask{\horidx}
\left[
    -
    \ClassWeight{\horidx}
    \RiskLabel{\horidx}
    \log \PredictedRisk{\horidx}
    -
    \left(
        1-\RiskLabel{\horidx}
    \right)
    \log
    \left(
        1-\PredictedRisk{\horidx}
    \right)
\right].
\label{eq:rce-loss}
\end{equation}

\subsection{Multi-Teacher Distillation}

Multi-horizon cancer risk prediction is commonly formulated as a multi-task learning problem, which may introduce optimization trade-offs between prediction horizons \citep{xin2022current}. To provide horizon-specialized supervision, \ours uses a set of independently trained teachers $\Teacher{1},\ldots,\Teacher{\numhorizons}$, with teachers sharing the embedding extraction backbone in common. One teacher is assigned to each prediction horizon, and for a given patient, each teacher processes the observed longitudinal visit-embedding sequence
$
\left\{
    \VisitEmbedding{\timeidx}
\right\}_{\timeidx\in\ObservedTimes}
$
and is optimized only for its corresponding cumulative prediction target. Specifically, teacher $\Teacher{\horidx}$ and the $\horidx$-th student output estimate the same cumulative binary event, namely whether cancer is diagnosed within horizon $\horidx$,
$\mathbb{P}\!\left(T_{\mathrm{cancer}} \le \horidx \mid \cdot \right),
$
while conditioning on different available information. Thus, independent teacher training does not changes the probabilistic interpretation of the teacher/ student outputs.

The teachers are first trained independently using the available longitudinal histories and valid labels for their assigned horizons. Their parameters are then frozen throughout student training. For horizon $\horidx$, let $\TeacherLogit{\horidx}$ denote the logit produced by teacher $\Teacher{\horidx}$ and $\StudentLogit{\horidx}$ the corresponding student logit. For distillation, we use the temperature-softened probabilities:
$
q_{\horidx}^{\teacher}
=
\sigma\!(\TeacherLogit{\horidx}/\beta)$, $
q_{\horidx}^{\student}
=
\sigma\!(
    \StudentLogit{\horidx}/\beta),
$
where $\beta>0$ is the distillation temperature. Importantly, the predictions across the $\numhorizons$ horizons do not form a categorical probability distribution and are not required to sum to one. Instead, each horizon defines a separate Bernoulli distribution for the cumulative event $T_{\mathrm{cancer}}\leq\horidx$. Accordingly, probability distillation is performed independently at each horizon:
\begin{equation}
\LogitKDLoss
=
\frac{
    1
}{
    \sum_{\horidx=1}^{\numhorizons}
    \LabelMask{\horidx}
}
\sum_{\horidx=1}^{\numhorizons}
\LabelMask{\horidx}
\,
D_{\mathrm{KL}}
\left(
    \operatorname{Bern}\!\left(q_{\horidx}^{\teacher}\right)
    \,\middle\|\,
    \operatorname{Bern}\!\left(q_{\horidx}^{\student}\right)
\right),
\label{eq:logit-distillation}
\end{equation}
where $\LabelMask{\horidx}\in\{0,1\}$ indicates whether the outcome at horizon $\horidx$ is known. The KL divergence therefore compares the two valid Bernoulli distributions
$
(q_{\horidx}^{\teacher},1-q_{\horidx}^{\teacher})
$
and
$
(q_{\horidx}^{\student},1-q_{\horidx}^{\student})
$
rather than the vectors of predictions across horizons. Each teacher therefore supervises only the student output corresponding to its assigned horizon, while the student jointly predicts the complete cumulative risk trajectory through its shared temporal representation and risk-prediction head.

\subsection{Overall Training Objective and Inference}

Training is performed in two stages. First, the teachers are trained using the available longitudinal screening histories and then frozen. The student is subsequently optimized using the combined objective
\begin{equation}
\mathcal{L}_{\mathrm{total}}
=
\mathcal{L}_{\mathrm{RCE}}
+
\lambda_f
\mathcal{L}_{\mathrm{KD}}^{\mathrm{feature}}
+
\lambda_l
\LogitKDLoss,
\label{eq:total-objective}
\end{equation}
where $\lambda_l$ and $\lambda_f$ control the contributions of distillation losses. During this stage, the image encoder, view aggregation module, and teachers remain frozen, while the prior embedding predictor, student temporal aggregator, and risk-prediction head are jointly optimized.

At inference time, the complete teacher pathway is discarded. Given only the current screening examination, the frozen visit representation modules produce the current visit embedding $\VisitEmbedding{0}$. The prior embedding predictor then estimates $\PredVisitEmbedding{-1},\ldots, \PredVisitEmbedding{-\maxprior}$, which are combined with the current embedding and processed by the student temporal aggregation backbone and additive hazard head. Therefore, deployment requires only the current mammographic examination and does not require access to true prior screening visits.
\section{Experimental Validation}
\label{sec:experiments}

\subsection{Experimental Setup}

\paragraph{Datasets and cohort construction.}
We evaluate \ours on three challenging longitudinal mammography cohorts: CSAW-CC \citep{ccc}, EMBED \citep{embed}, OMI-DB \citep{omidb}, all freely available for non-commercial research purposes subject to its corresponding agreement. Table~\ref{tab:dataset-summary} summarizes the cohort details. CSAW-CC is derived from the Cohort of Screen-Aged Women \citep{Dembrower2019} and is the same cohort used in LoMaR \citep{lomar} and VMRA \citep{vmra} evaluations. EMBED is a large and demographically diverse dataset containing mammographic examinations with linked clinical outcomes. OMI-DB is a longitudinal UK breast-screening database containing images, dates, and diagnosis records. For each dataset, examinations from the same patient are ordered chronologically and restricted to the four standard mammographic views. Each eligible examination is treated as a potential current visit and paired with preceding screening examinations. Examinations associated with malignancy at the current visit are excluded from the future-risk prediction cohort. Detailed analysis of the datasets is provided in Appendix~\ref{app:cohort-details}.
\begin{table}[H]
\centering
\caption{
Summary of the datasets used for the experiments. Max horizon shows max year distance between first and last exam. Number of Future malignant patients shows ones that have at least one malignant exam occurring after their first examination.
}
\label{tab:dataset-summary}
\small
\setlength{\linewidth}{6pt}
\begin{tabular}{lrrrrrr}
\toprule
Dataset
& \# Patients
& \# Exams
& \makecell{Max. \\ exams}
& \makecell{Max. \\ horizon}
& \makecell{\# Future \\ malignant patients} \\
\midrule
CSAW-CC, \cite{ccc}
& 8,723
& 24,694
& 6
& 8
& 533 \\

EMBED, \cite{embed}
& 16,487
& 42,369
& 9
& 7
& 402 \\

OMI-DB, \cite{omidb}
& 11,531
& 36,859
& 9
& 13
& 2,583 \\
\bottomrule
\end{tabular}
\end{table}
\textbf{Data splits.}
All sets are created at the patient level to prevent examinations from the same individual from appearing in multiple subsets. For splitting, we follow prior work \citep{lomar,vmra} and randomly assign 80\% of patients to training and 20\% to testing, with 25\% of the training patients used for validation. This procedure is repeated over 10 independent patient-level splits. We follow the evaluation protocol of LoMaR \citep{lomar}. For each prediction horizon, one eligible examination is sampled per test patient to prevent patients with longer screening histories from contributing disproportionately. Sampling is repeated 100 times within each split, meaning that for each training split, 100 tests are performed.

\textbf{Evaluation protocol and metrics.}
We report AUC and standardized partial AUC over false-positive rate (FPR) interval $[0,0.1]$ (pAUC@10). The pAUC emphasizes the clinically relevant low-FPR region, where excessive false positives may lead to unnecessary follow-up imaging or biopsy. The reported mean and standard deviation are then computed across the 10 patient-level splits (which are themselves averaged over 100 tests per set). Only examinations with a known outcome at the corresponding horizon are included.

\textbf{Models and implementation details.}
Teachers are first trained using the available observed longitudinal histories and are subsequently frozen during student optimization. $\HistoryPredictor(\cdot)$ is implemented as a three-layer fully connected network with ReLU activations and dropout of 0.1, and is also first pre-trained to reconstruct observed prior-visit embeddings from the current visit representation using the feature-level objective. Its parameters are then used to initialize the student. All hyperparameters, including backbone-specific configurations and distillation parameters, are tuned based on validation performance. All models are trained using Adam with an initial learning rate of $10^{-3}$ and cosine learning-rate scheduling for up to 30 epochs, with early stopping after 5 epochs without improvement in the validation objective.

\textbf{Baselines and Comparisons.} We compare \ours against single-exam and longitudinal mammography risk prediction baselines under different inference history-availability settings. First, we include single-exam state-of-the-art Mirai \cite{Yala2021}, trained and evaluated using only the current screening examination. We also evaluate transformer-based and VMamba-based history aggregation LoMaR \cite{lomar} and VMRA \cite{vmra}, both trained using the available longitudinal screening histories. For LoMaR and VMRA, we consider two inference settings. In the full-history setting, denoted by $\eta =4$, the models are trained with longitudinal history and receive up to four observed prior examinations at inference. 

This setting serves as the full-history reference for performance when historical examinations are available at deployment. In the no-history setting, denoted by $\eta =0$, the same longitudinally trained models are evaluated after all prior examinations are removed, such that only the current examination is provided at inference. The difference between $\eta =4$ and $\eta =0$ therefore quantifies the performance loss caused by unavailable screening history at deployment.
LoMaR+\ours and VMRA+\ours are evaluated under the same $\eta =0$ inference constraint. They do not access true prior examinations at test time. Instead, the student combines the current visit embedding with latent historical embeddings predicted from the current examination. The primary comparison is therefore between each \ours model and its corresponding longitudinally trained $\eta =0$ baseline, since both use only the current examination at deployment, while the $\eta =4$ models provide full-history reference performance.

\subsection{Results and Discussion}

\textbf{Longitudinal history primarily benefits longer-horizon risk prediction.}
Table~\ref{tab:main} first checks the effect of removing observed screening history at inference. Across both LoMaR and VMRA, the performance difference between full-history ($\eta =4$) and no-history ($\eta =0$) inference is generally small at shorter horizons but becomes more pronounced toward 4--5 years. This pattern is observed across the three cohorts and is particularly clear in pAUC@10\%. For example, on CSAW-CC, 5-year pAUC decreases from 0.740 to 0.711 for LoMaR and from 0.745 to 0.710 for VMRA when prior examinations are removed. These results show that longitudinally trained models remain sensitive to the availability of true screening history at deployment, with longitudinal information contributing most strongly to longer-term risk estimation.

\textbf{Privileged history supervision recovers much of the missing-history gap.}
Under the same current-exam-only inference constraint ($\eta =0$), \ours consistently improves the corresponding LoMaR and VMRA baselines at the horizons where removing history causes the largest degradation. At 5 years, LoMaR+\ours and VMRA+\ours improve AUC/pAUC from 0.829/0.711 and 0.829/0.710 to 0.853/0.752 and 0.855/0.757 on CSAW-CC, from 0.759/0.631 and 0.759/0.631 to 0.793/0.683 and 0.794/0.686 on OMI-DB, and from 0.768/0.641 and 0.769/0.646 to 0.786/0.650 and 0.791/0.674 on EMBED. Overall, the effect is strongest at 4--5 years and is observed with both temporal backbones, indicating that the benefit is not specific to either the transformer or recurrent aggregation architecture. Compared with the single-exam Mirai baseline, the same pattern is observed: the clearest advantages emerge at longer horizons. \ours also approaches the corresponding full-history references despite receiving no true prior examinations at inference, although the degree of recovery varies across datasets.


\colorlet{lightorchid}{Orchid!10}
\begin{table*}[t]
\centering
\scriptsize
\setlength{\tabcolsep}{3.2pt}
\renewcommand{\arraystretch}{1.05}

\caption{
Mean$\pm$std AUC and pAUC@10\% for 1-5 year cancer risk prediction on CSAW-CC, OMIDB, and EMBED. Inference history \cmark is $\eta =4$ and \xmark is $\eta =0$. \ours has no true prior at inference.
}
\label{tab:main}

\resizebox{0.98\textwidth}{!}{
\begin{tabular}{c c l c c c c c c}
\toprule
Data & Met. & Model & Inference Hist. & 1y & 2y & 3y & 4y & 5y \\
\midrule


\multirow[c]{14}{*}{
    \rotatebox[origin=c]{90}{\strut CSAW-CC}
}
&
\multirow[c]{7}{*}{
    \rotatebox[origin=c]{90}{\strut Full AUC}
}
& LoMaR (MICCAI,24) & \cmark
& \std{0.914}{0.023}
& \std{0.865}{0.020}
& \std{0.851}{0.017}
& \std{0.841}{0.019}
& \std{0.851}{0.016} \\

&
& VMRA (MICCAI,25) & \cmark
& \std{0.920}{0.019}
& \std{0.868}{0.020}
& \std{0.851}{0.017}
& \std{0.842}{0.017}
& \std{0.851}{0.017} \\

\cmidrule(lr){3-9}

&
& Mirai & \xmark
& \textbf{\std{0.924}{0.020}}
& \std{0.872}{0.016}
& \textbf{\std{0.853}{0.015}}
& \std{0.837}{0.014}
& \std{0.829}{0.015} \\

&
& LoMaR (MICCAI,24) & \xmark
& \std{0.922}{0.020}
& \textbf{\std{0.873}{0.016}}
& \textbf{\std{0.853}{0.015}}
& \std{0.837}{0.014}
& \std{0.829}{0.015} \\

&
& VMRA (MICCAI,25) & \xmark
& \std{0.922}{0.020}
& \std{0.872}{0.017}
& \textbf{\std{0.853}{0.016}}
& \std{0.836}{0.015}
& \std{0.829}{0.015} \\

&
& \hlcell{LoMaR+\ours}
& \hlcell{\xmark}
& \hlcell{\std{0.913}{0.022}}
& \hlcell{\std{0.865}{0.019}}
& \hlcell{\std{0.852}{0.016}}
& \hlcell{\textbf{\std{0.845}{0.015}}}
& \hlcell{\textbf{\std{0.853}{0.015}}} \\

&
& \hlcell{VMRA+\ours}
& \hlcell{\xmark}
& \hlcell{\std{0.920}{0.018}}
& \hlcell{\std{0.869}{0.018}}
& \hlcell{\std{0.852}{0.016}}
& \hlcell{\textbf{\std{0.847}{0.015}}}
& \hlcell{\textbf{\std{0.855}{0.017}}} \\


\cmidrule(lr){2-9}

&
\multirow[c]{7}{*}{
    \rotatebox[origin=c]{90}{\strut pAUC}
}
& LoMaR (MICCAI,24) & \cmark
& \std{0.817}{0.023}
& \std{0.749}{0.020}
& \std{0.738}{0.018}
& \std{0.731}{0.018}
& \std{0.740}{0.018} \\

&
& VMRA (MICCAI,25) & \cmark
& \std{0.822}{0.020}
& \std{0.752}{0.020}
& \std{0.736}{0.018}
& \std{0.728}{0.019}
& \std{0.745}{0.021} \\

\cmidrule(lr){3-9}

&
& Mirai & \xmark
& \textbf{\std{0.824}{0.023}}
& \std{0.753}{0.019}
& \textbf{\std{0.735}{0.018}}
& \std{0.715}{0.018}
& \std{0.711}{0.021} \\

&
& LoMaR (MICCAI,24) & \xmark
& \textbf{\std{0.824}{0.023}}
& \textbf{\std{0.754}{0.019}}
& \textbf{\std{0.735}{0.018}}
& \std{0.715}{0.018}
& \std{0.711}{0.020} \\

&
& VMRA (MICCAI,25) & \xmark
& \std{0.823}{0.023}
& \std{0.752}{0.019}
& \std{0.734}{0.018}
& \std{0.714}{0.019}
& \std{0.710}{0.021} \\

&
& \hlcell{LoMaR+\ours}
& \hlcell{\xmark}
& \hlcell{\std{0.810}{0.031}}
& \hlcell{\std{0.744}{0.024}}
& \hlcell{\std{0.734}{0.015}}
& \hlcell{\textbf{\std{0.735}{0.020}}}
& \hlcell{\textbf{\std{0.752}{0.019}}} \\

&
& \hlcell{VMRA+\ours}
& \hlcell{\xmark}
& \hlcell{\std{0.818}{0.020}}
& \hlcell{\std{0.749}{0.017}}
& \hlcell{\std{0.733}{0.017}}
& \hlcell{\textbf{\std{0.734}{0.017}}}
& \hlcell{\textbf{\std{0.757}{0.018}}} \\

\midrule


\multirow[c]{14}{*}{
    \rotatebox[origin=c]{90}{\strut OMIDB}
}
&
\multirow[c]{7}{*}{
    \rotatebox[origin=c]{90}{\strut Full AUC}
}
& LoMaR (MICCAI,24) & \cmark
& \std{0.900}{0.008}
& \textbf{\std{0.891}{0.008}}
& \std{0.786}{0.006}
& \std{0.797}{0.007}
& \std{0.797}{0.007} \\

&
& VMRA (MICCAI,25) & \cmark
& \std{0.900}{0.008}
& \std{0.890}{0.008}
& \std{0.788}{0.009}
& \std{0.795}{0.009}
& \std{0.795}{0.009} \\

\cmidrule(lr){3-9}

&
& Mirai & \xmark
& \std{0.844}{0.011}
& \std{0.838}{0.011}
& \std{0.765}{0.007}
& \std{0.761}{0.009}
& \std{0.760}{0.008} \\

&
& LoMaR (MICCAI,24) & \xmark
& \std{0.845}{0.011}
& \std{0.839}{0.011}
& \std{0.767}{0.007}
& \std{0.760}{0.009}
& \std{0.759}{0.008} \\

&
& VMRA (MICCAI,25) & \xmark
& \std{0.846}{0.010}
& \std{0.840}{0.011}
& \std{0.765}{0.008}
& \std{0.759}{0.009}
& \std{0.759}{0.009} \\

&
& \hlcell{LoMaR+\ours}
& \hlcell{\xmark}
& \hlcell{\textbf{\std{0.893}{0.009}}}
& \hlcell{\textbf{\std{0.884}{0.009}}}
& \hlcell{\textbf{\std{0.781}{0.006}}}
& \hlcell{\textbf{\std{0.793}{0.008}}}
& \hlcell{\textbf{\std{0.793}{0.007}}} \\

&
& \hlcell{VMRA+\ours}
& \hlcell{\xmark}
& \hlcell{\textbf{\std{0.896}{0.007}}}
& \hlcell{\textbf{\std{0.887}{0.009}}}
& \hlcell{\textbf{\std{0.785}{0.009}}}
& \hlcell{\textbf{\std{0.794}{0.009}}}
& \hlcell{\textbf{\std{0.794}{0.008}}} \\


\cmidrule(lr){2-9}

&
\multirow[c]{7}{*}{
    \rotatebox[origin=c]{90}{\strut pAUC}
}
& LoMaR (MICCAI,24) & \cmark
& \std{0.760}{0.008}
& \std{0.754}{0.008}
& \std{0.671}{0.005}
& \std{0.690}{0.007}
& \std{0.690}{0.007} \\

&
& VMRA (MICCAI,25) & \cmark
& \std{0.759}{0.009}
& \std{0.755}{0.009}
& \std{0.670}{0.006}
& \std{0.689}{0.008}
& \std{0.690}{0.007} \\

\cmidrule(lr){3-9}

&
& Mirai & \xmark
& \std{0.699}{0.008}
& \std{0.694}{0.008}
& \std{0.631}{0.006}
& \std{0.631}{0.007}
& \std{0.632}{0.006} \\

&
& LoMaR (MICCAI,24) & \xmark
& \std{0.704}{0.009}
& \std{0.699}{0.009}
& \std{0.634}{0.007}
& \std{0.631}{0.007}
& \std{0.631}{0.006} \\

&
& VMRA (MICCAI,25) & \xmark
& \std{0.706}{0.008}
& \std{0.701}{0.008}
& \std{0.632}{0.008}
& \std{0.631}{0.007}
& \std{0.631}{0.007} \\

&
& \hlcell{LoMaR+\ours}
& \hlcell{\xmark}
& \hlcell{\textbf{\textbf{\std{0.743}{0.011}}}}
& \hlcell{\textbf{\std{0.739}{0.011}}}
& \hlcell{\textbf{\std{0.663}{0.006}}}
& \hlcell{\textbf{\std{0.683}{0.006}}}
& \hlcell{\textbf{\std{0.683}{0.006}}} \\

&
& \hlcell{VMRA+\ours}
& \hlcell{\xmark}
& \hlcell{\textbf{\std{0.746}{0.010}}}
& \hlcell{\textbf{\std{0.743}{0.010}}}
& \hlcell{\textbf{\std{0.665}{0.006}}}
& \hlcell{\textbf{\std{0.684}{0.006}}}
& \hlcell{\textbf{\std{0.686}{0.007}}} \\

\midrule


\multirow[c]{14}{*}{
    \rotatebox[origin=c]{90}{\strut EMBED}
}
&
\multirow[c]{7}{*}{
    \rotatebox[origin=c]{90}{\strut Full AUC}
}
& LoMaR (MICCAI,24) & \cmark
& \std{0.797}{0.043}
& \std{0.770}{0.034}
& \std{0.773}{0.027}
& \std{0.775}{0.028}
& \std{0.784}{0.026} \\

&
& VMRA (MICCAI,25) & \cmark
& \std{0.805}{0.036}
& \std{0.772}{0.030}
& \std{0.768}{0.025}
& \std{0.775}{0.023}
& \std{0.788}{0.028} \\

\cmidrule(lr){3-9}

&
& Mirai & \xmark
& \std{0.806}{0.041}
& \std{0.775}{0.033}
& \std{0.769}{0.030}
& \std{0.768}{0.029}
& \std{0.768}{0.031} \\

&
& LoMaR (MICCAI,24) & \xmark
& \std{0.809}{0.041}
& \std{0.778}{0.033}
& \std{0.771}{0.031}
& \std{0.767}{0.031}
& \std{0.768}{0.032} \\

&
& VMRA (MICCAI,25) & \xmark
& \textbf{\std{0.812}{0.039}}
& \textbf{\std{0.780}{0.031}}
& \std{0.772}{0.028}
& \std{0.768}{0.027}
& \std{0.769}{0.028} \\

&
& \hlcell{LoMaR+\ours}
& \hlcell{\xmark}
& \hlcell{\std{0.800}{0.042}}
& \hlcell{\std{0.774}{0.033}}
& \hlcell{\textbf{\std{0.776}{0.027}}}
& \hlcell{\textbf{\std{0.781}{0.026}}}
& \hlcell{\textbf{\std{0.786}{0.022}}} \\

&
& \hlcell{VMRA+\ours}
& \hlcell{\xmark}
& \hlcell{\std{0.803}{0.037}}
& \hlcell{\std{0.771}{0.031}}
& \hlcell{\std{0.770}{0.024}}
& \hlcell{\textbf{\std{0.779}{0.023}}}
& \hlcell{\textbf{\std{0.791}{0.028}}} \\


\cmidrule(lr){2-9}

&
\multirow[c]{7}{*}{
    \rotatebox[origin=c]{90}{\strut pAUC}
}
& LoMaR (MICCAI,24) & \cmark
& \std{0.672}{0.035}
& \std{0.642}{0.028}
& \std{0.644}{0.026}
& \std{0.653}{0.024}
& \std{0.649}{0.025} \\

&
& VMRA (MICCAI,25) & \cmark
& \std{0.689}{0.051}
& \std{0.648}{0.037}
& \std{0.644}{0.032}
& \std{0.660}{0.029}
& \std{0.671}{0.034} \\

\cmidrule(lr){3-9}

&
& Mirai & \xmark
& \std{0.680}{0.035}
& \std{0.646}{0.029}
& \std{0.640}{0.029}
& \std{0.646}{0.029}
& \std{0.638}{0.030} \\

&
& LoMaR (MICCAI,24) & \xmark
& \std{0.688}{0.035}
& \std{0.653}{0.030}
& \std{0.646}{0.028}
& \std{0.648}{0.029}
& \std{0.641}{0.031} \\

&
& VMRA (MICCAI,25) & \xmark
& \std{0.694}{0.037}
& \std{0.654}{0.030}
& \std{0.648}{0.029}
& \std{0.652}{0.030}
& \std{0.646}{0.034} \\

&
& \hlcell{LoMaR+\ours}
& \hlcell{\xmark}
& \hlcell{\std{0.676}{0.033}}
& \hlcell{\std{0.648}{0.027}}
& \hlcell{\textbf{\std{0.651}{0.027}}}
& \hlcell{\textbf{\std{0.660}{0.027}}}
& \hlcell{\textbf{\std{0.650}{0.025}}} \\

&
& \hlcell{VMRA+\ours}
& \hlcell{\xmark}
& \hlcell{\textbf{\std{0.694}{0.040}}}
& \hlcell{\textbf{\std{0.656}{0.031}}}
& \hlcell{\textbf{\std{0.649}{0.027}}}
& \hlcell{\textbf{\std{0.665}{0.026}}}
& \hlcell{\textbf{\std{0.674}{0.031}}} \\
\bottomrule
\end{tabular}
}
\end{table*}
\begin{table*}[t]
\centering
\scriptsize
\renewcommand{\arraystretch}{1.10}

\caption{
$t$-test $p$-values comparing each missing-history baseline ($\eta =0$) with its corresponding \ours model across 10 splits. Bold values indicate statistical significance at $p<0.05$.
}
\label{tab:auc-significance}

\begin{tabular}{l c c c c c c c c c c}
\toprule
& \multicolumn{5}{c}{\makecell{LoMaR+\ours \\ vs.\ LoMaR ($\eta =0$)}}
& \multicolumn{5}{c}{\makecell{VMRA+\ours \\ vs.\ VMRA ($\eta =0$)}} \\
\cmidrule(lr){2-6}
\cmidrule(lr){7-11}

Dataset
& 1y & 2y & 3y & 4y & 5y
& 1y & 2y & 3y & 4y & 5y \\
\midrule

CSAW-CC
& \textbf{0.0045}
& \textbf{0.0175}
& 0.2806
& 0.1337
& \textbf{1.0e-07}
& \textbf{0.0110}
& \textbf{0.0341}
& 0.1584
& \textbf{6.3e-06}
& \textbf{6.3e-09} \\

OMI-DB
& \textbf{1.4e-10}
& \textbf{1.4e-10}
& \textbf{5.9e-07}
& \textbf{5.2e-10}
& \textbf{3.8e-10}
& \textbf{1.2e-10}
& \textbf{3.0e-10}
& \textbf{9.3e-07}
& \textbf{5.2e-10}
& \textbf{2.4e-10} \\

EMBED
& \textbf{0.0088}
& 0.2001
& 0.0520
& \textbf{0.0034}
& \textbf{0.0067}
& \textbf{0.0449}
& \textbf{0.0494}
& 0.7111
& \textbf{0.0114}
& \textbf{0.0006} \\

\bottomrule
\end{tabular}
\end{table*}

\textbf{The long-horizon improvements are statistically significant.}
The statistical comparisons in Table~\ref{tab:auc-significance} support the same horizon-dependent pattern. At the 5-year horizon, \ours significantly improves over its corresponding no-history baseline for both LoMaR and VMRA on all three datasets. OMI-DB, which has the richest available history in the cohort, shows significant improvements across all five horizons, whereas significance is more selective at shorter horizons on CSAW-CC and EMBED.


\begin{wrapfigure}{r}{0.53\columnwidth}
    \centering
    \vspace{-8pt}
    \includegraphics[width=0.53\textwidth]{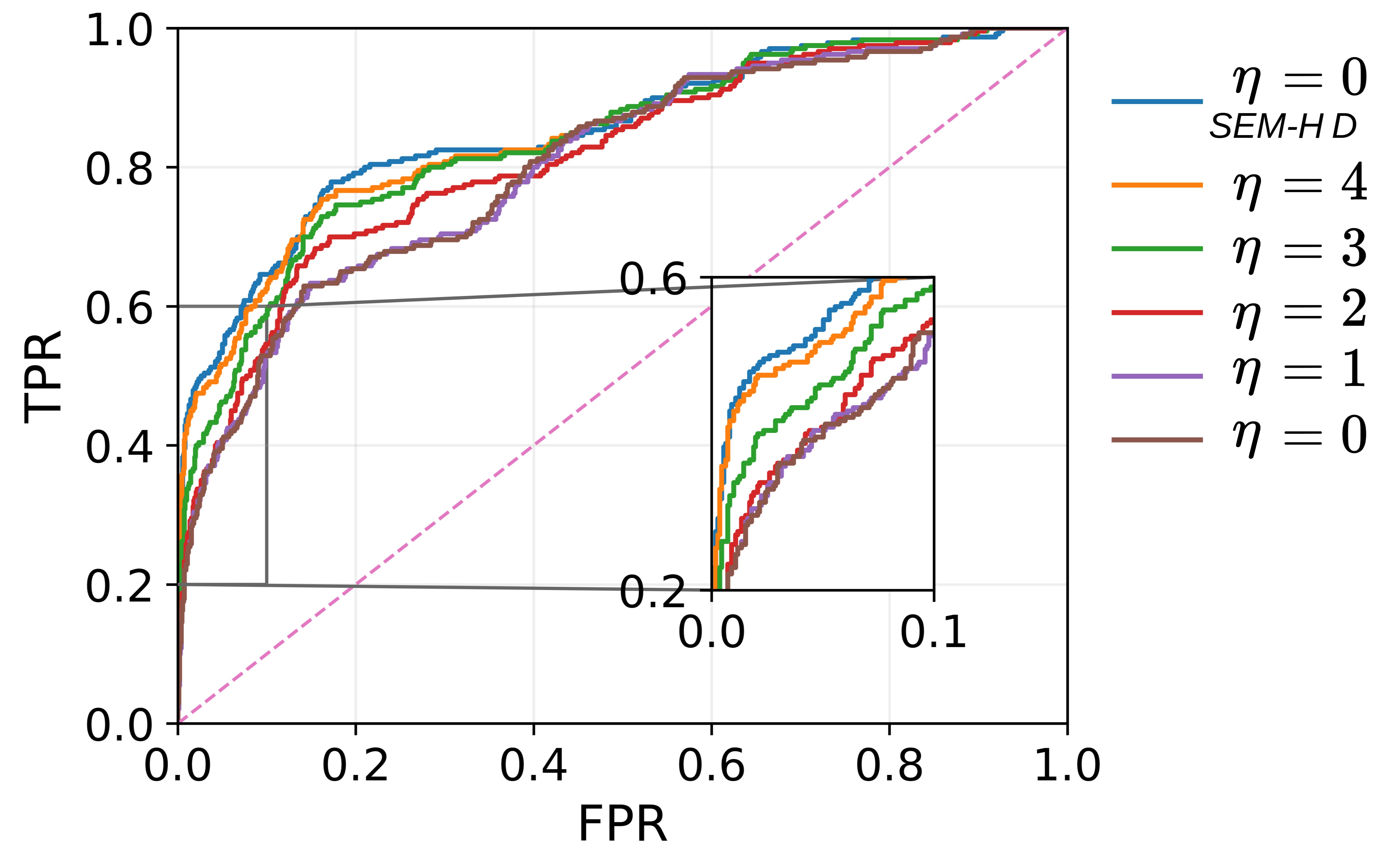}
    \caption{ROC curves compare \ours (VMRA/ CSAW) under different history-availability settings for CSAW-CC. \ours has $\eta =0$, yet improves sensitivity in the low-FPR region compared with no-history baselines and approaches or exceeds full-history models.}
    \label{fig:roc_vmra}
    \vspace{-8pt}
\end{wrapfigure}
\textbf{Low-FPR Region Sensitivity.} 
The pAUC improvements are also visible directly in the ROC operating characteristics (Fig.~\ref{fig:roc_vmra}). Under current-exam-only inference, removing history reduces sensitivity at low false-positive rates, whereas \ours shifts the ROC curve toward the full-history reference despite having no access to true prior examinations. This suggests that privileged longitudinal supervision is not only improving global ranking performance, but is particularly useful in the operating region emphasized by pAUC@10\%, where increases in false positives are costly.

\begin{figure}[t]
    \centering
    \includegraphics[width=\linewidth]{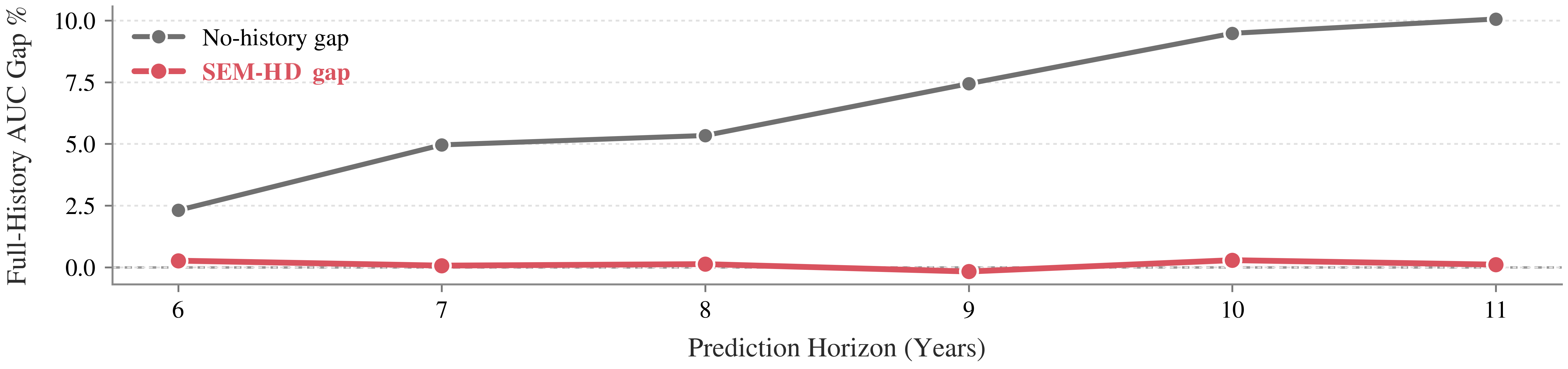}
    \caption{Long-horizon analysis on OMI-DB and LoMaR. We report the AUC gap relative to full-history inference, computed at each prediction horizon as $\mathrm{AUC}_{\mathrm{Full\ History}}-\mathrm{AUC}_{\mathrm{model}}$. The no-history inference increasingly diverges from the full-history reference as the prediction horizon extends from 6 to 11 years, whereas \ours remains consistently close to the full-history performance. }
    \label{fig:extended-horizon-gap}
\end{figure}
\textbf{The benefit of privileged history becomes more pronounced as the prediction horizon extends.} The extended OMI-DB analysis in Fig.~\ref{fig:extended-horizon-gap} reveals a complementary pattern beyond the standard 5-year window. As the prediction horizon increases from 6 to 11 years, the no-history model progressively diverges from the full-history reference, indicating increasing dependence on longitudinal information. In contrast, \ours remains close to the full-history model across the same horizons. The no-history AUC gap grows from approximately 2.3 percentage points at 6 years to 10.0 points at 11 years, whereas \ours remains within approximately 0.5 points of the full-history reference across the evaluated horizons. Together with the 1--5 year results, this indicates that the value of privileged temporal supervision increases as the prediction task becomes more dependent on historical context.

\textbf{The gains cannot be explained by simply populating missing temporal positions.}
We next test whether \ours benefits merely from presenting the longitudinal backbone with a complete sequence of embeddings rather than from learning patient-specific historical information. 

\begin{wrapfigure}{r}{0.53\columnwidth}
    \centering
    \includegraphics[width=\linewidth]{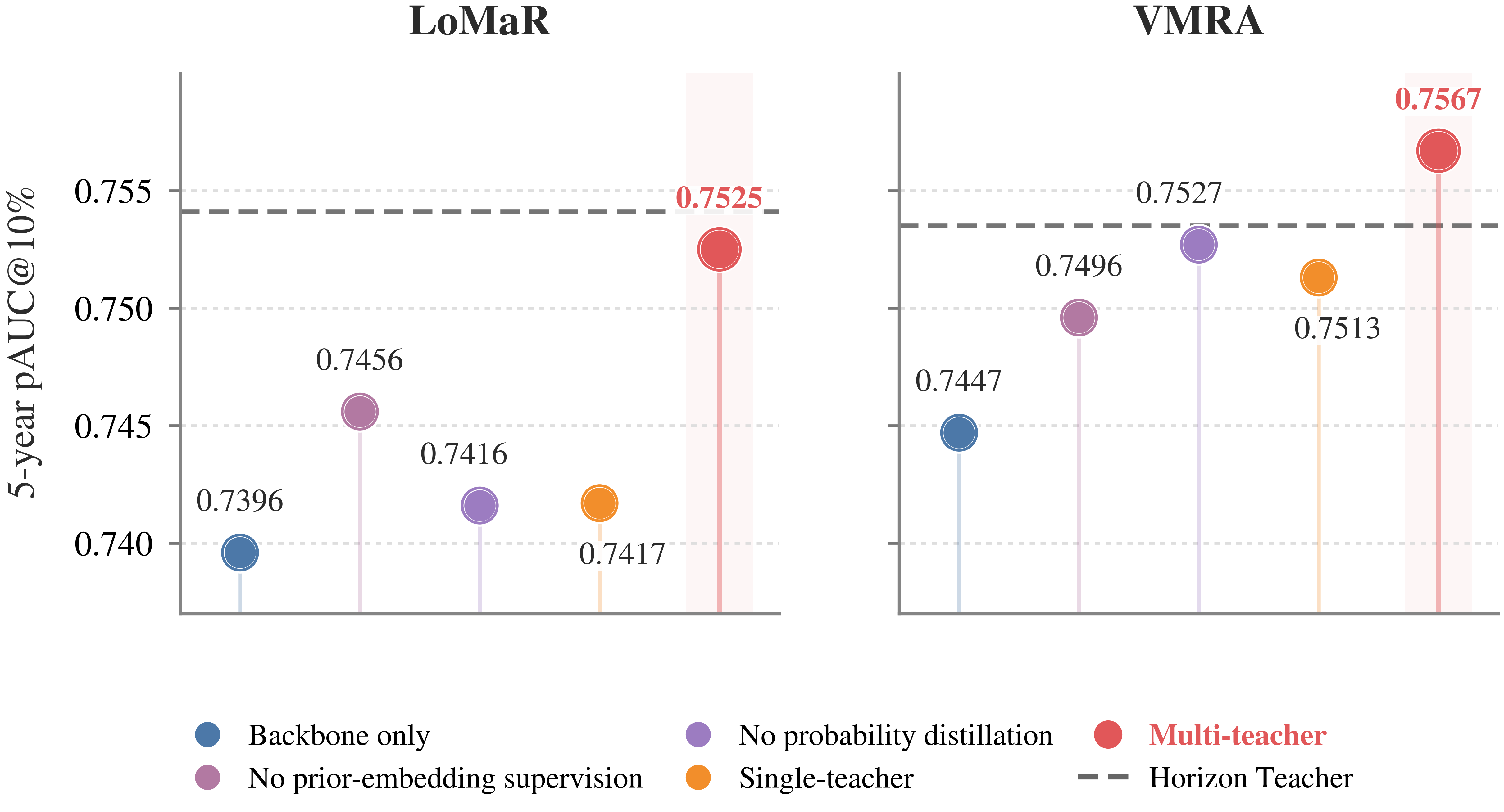}
    \caption{Ablation of \ours at 5-year horizon on CSAW-CC using LoMaR and VMRA backbones. \textit{Backbone only} uses original supervised risk objective without \ours components. \textit{No prior-embedding supervision} removes the prior-visit embedding predictor training component from \ours, while \textit{No probability distillation} removes probability distillation from \ours. \textit{Single-teacher} uses both components with a single teacher, and \textit{Multi-teacher} is the complete \ours model. The dashed line shows the corresponding 5-year Horizon Teacher performance. The complete \ours achieves the highest deployable student pAUC for both backbones.}
    \label{fig:phd_abblation}
    \vspace{-8pt}
\end{wrapfigure} 
We compare against masked training (where VMRA is trained with randomly masked historical visits to improve robustness to missing history at inference), replacement of each missing visit by the current embedding (All $x^0$), and replacement by 
the per year mean historical embedding computed from the training set (All Mean). None of these alternatives reproduces the long-horizon improvement of \ours. At 4--5 years, \ours clearly outperforms both the no-history model and heuristic filling strategies, whereas the alternatives provide little or no improvement. This indicates that the gain is not explained by the presence of additional temporal tokens alone, it depends on learning patient-specific substitutes for the unavailable history.
\begin{figure*}
    \centering
    \includegraphics[width=0.9\textwidth]{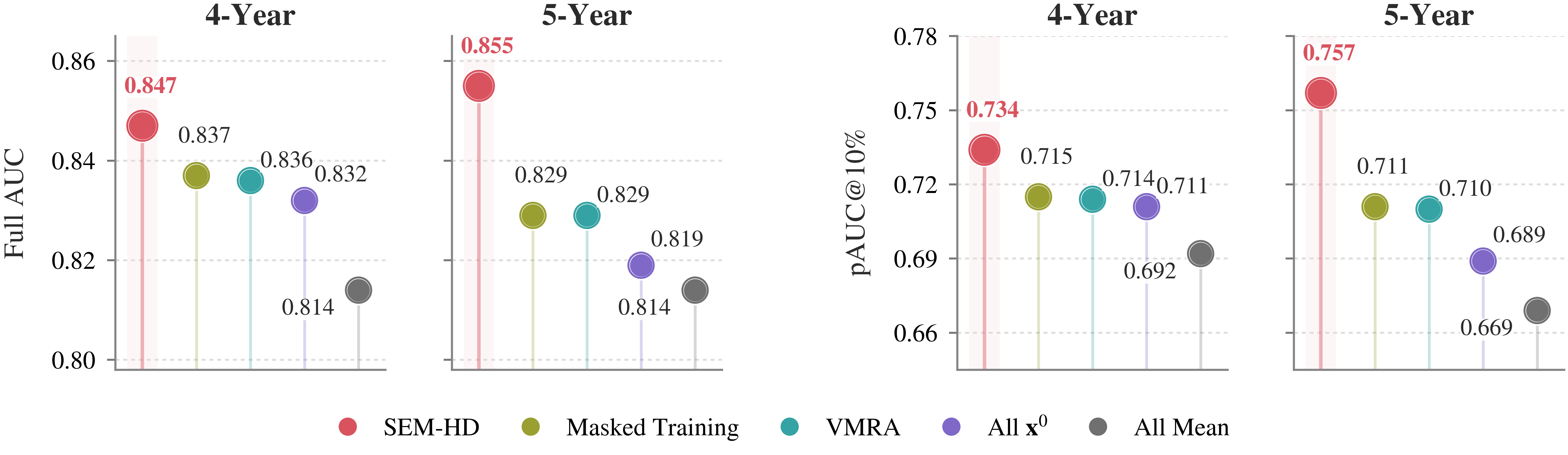}
    \caption{Comparison against alternative missing-history handling strategies for VMRA and CSAW-CC under no-history inference. All methods operate without access to true prior mammograms at test time.}
    \label{fig:missing}
\end{figure*} 

\textbf{Ablation study.}
Figure~\ref{fig:phd_abblation} examines the contribution of the main \ours components at the 5-year horizon. Removing either prior-embedding supervision or probability distillation reduces the improvement over the backbone-only model, indicating that latent history prediction and teacher-based risk supervision provide complementary benefits. Using both components with a single teacher remains competitive, while the complete multi-teacher \ours achieves the highest deployable pAUC for both LoMaR and VMRA, reaching 0.7525 and 0.7567, respectively. The resulting students closely approach the corresponding 5-year horizon teachers, with the VMRA student slightly exceeding its teacher. Overall, these results indicate that the main benefit comes from combining latent-history prediction with distilled risk supervision, while horizon-specific teacher specialization provides an additional improvement. 

Taken together, these findings suggest that privileged temporal supervision is most useful when the downstream task strongly depends on historical context, and that preserving the longitudinal input structure is more effective than heuristic substitution of missing history. Additional analyses of the historical embedding predictor, computational cost, complete statistical comparisons, single- versus multi-teacher distillation, and hyperparameter sensitivity are provided in Appendices~\ref{app:missing-history-results}--~\ref{app:hp}.

\section{Conclusion and Limitations}  

We introduced \ours for breast cancer risk prediction when prior examinations are unavailable at deployment. \ours uses longitudinal history only during training, combining latent prior prediction with horizon-specific multi-teacher distillation, while inference requires only the current examination. Across three cohorts and two temporal backbones, \ours consistently improved long-horizon performance over no-history longitudinal models and recovered much of the gap to full-history inference, particularly at low false-positive rates. More broadly, these results suggest that unavailable temporal context can be treated as privileged supervision while preserving the underlying longitudinal modeling structure. \ours nevertheless has additional teacher-training cost, and its behavior with partially available history at inference remains to be studied. Future work will investigate more efficient teacher formulations, flexible temporal prediction, and broader prospective validation.

\bibliography{main}

@ARTICLE{Wilkinson2022-dg,
  title     = "Understanding breast cancer as a global health concern",
  author    = "Wilkinson, Louise and Gathani, Toral",
  journal   = "Br. J. Radiol.",
  publisher = "Oxford University Press (OUP)",
  volume    =  95,
  number    =  1130,
  pages     = "20211033",
  month     =  feb,
  year      =  2022,
  language  = "en"
}

@article{hinton2015,
  title={Distilling the knowledge in a neural network},
  author={Hinton, Geoffrey and Vinyals, Oriol and Dean, Jeff},
  journal={arXiv preprint arXiv:1503.02531},
  year={2015}
}

@ARTICLE{Hakama2008-fi,
  title     = "Cancer screening: evidence and practice in Europe 2008",
  author    = "Hakama, Matti and Coleman, Michel P and Alexe, Delia-Marina and
               Auvinen, Anssi",
  journal   = "Eur. J. Cancer",
  publisher = "Elsevier BV",
  volume    =  44,
  number    =  10,
  pages     = "1404--1413",
  month     =  jul,
  year      =  2008,
  language  = "en"
}

@ARTICLE{Bray2024-sv,
  title     = "Global cancer statistics 2022: {GLOBOCAN} estimates of incidence
               and mortality worldwide for 36 cancers in 185 countries",
  author    = "Bray, Freddie and Laversanne, Mathieu and Sung, Hyuna and
               Ferlay, Jacques and Siegel, Rebecca L and Soerjomataram,
               Isabelle and Jemal, Ahmedin",
  journal   = "CA Cancer J. Clin.",
  publisher = "Wiley",
  volume    =  74,
  number    =  3,
  pages     = "229--263",
  month     =  may,
  year      =  2024,
  copyright = "http://creativecommons.org/licenses/by-nc-nd/4.0/",
  language  = "en"
}

@ARTICLE{Wilkinson2025-nx,
  title    = "Cost-effectiveness of breast cancer screening using digital
              mammography in Canada",
  author   = "Wilkinson, Anna N and Mainprize, James G and Yaffe, Martin J and
              Robinson, Jessica and Cordeiro, Erin and Look Hong, Nicole J and
              Williams, Phillip and Moideen, Nikitha and Renaud, Julie and
              Seely, Jean M and Rushton, Moira",
  journal  = "JAMA Netw. Open",
  volume   =  8,
  number   =  1,
  pages    = "e2452821",
  month    =  jan,
  year     =  2025,
  language = "en"
}

@ARTICLE{Kim2021-cy,
  title     = "Assessing risk of breast cancer: A review of risk prediction
               models",
  author    = "Kim, Geunwon and Bahl, Manisha",
  journal   = "J. Breast Imaging",
  publisher = "Oxford University Press (OUP)",
  volume    =  3,
  number    =  2,
  pages     = "144--155",
  month     =  mar,
  year      =  2021,
  copyright = "https://academic.oup.com/journals/pages/open\_access/funder\_policies/chorus/standard\_publication\_model",
  language  = "en"
}

@inproceedings{thrun2025reconsidering,
  title={Reconsidering Explicit Longitudinal Mammography Alignment for Enhanced Breast Cancer Risk Prediction},
  author={Thrun, Solveig and Hansen, Stine and Sun, Zijun and Blum, Nele and Salahuddin, Suaiba A and Wickstr{\o}m, Kristoffer and Wetzer, Elisabeth and Jenssen, Robert and Stille, Maik and Kampffmeyer, Michael},
  booktitle={International Conference on Medical Image Computing and Computer-Assisted Intervention},
  pages={495--505},
  year={2025},
  organization={Springer}
}

@ARTICLE{Mendes2025-br,
  title     = "Deep learning in breast cancer risk prediction: a review of
               recent applications in full-field digital mammography",
  author    = "Mendes, Jo{\~a}o and Oliveira, Bernardo and Ara{\'u}jo, Carolina
               and Galr{\~a}o, Joana and Mota, Ana M and Garcia, Nuno C and
               Matela, Nuno",
  journal   = "Front. Oncol.",
  publisher = "Frontiers Media SA",
  volume    =  15,
  number    =  1656842,
  pages     = "1656842",
  month     =  sep,
  year      =  2025,
  copyright = "https://creativecommons.org/licenses/by/4.0/",
  language  = "en"
}

@ARTICLE{Kim2023-ig,
  title     = "Deep learning analysis of mammography for breast cancer risk
               prediction in Asian women",
  author    = "Kim, Hayoung and Lim, Jihe and Kim, Hyug-Gi and Lim, Yunji and
               Seo, Bo Kyoung and Bae, Min Sun",
  journal   = "Diagnostics (Basel)",
  publisher = "MDPI AG",
  volume    =  13,
  number    =  13,
  pages     = "2247",
  month     =  jul,
  year      =  2023,
  copyright = "https://creativecommons.org/licenses/by/4.0/",
  language  = "en"
}

@ARTICLE{Santeramo2024-zx,
  title     = "Are better {AI} algorithms for breast cancer detection also
               better at predicting risk? A paired case-control study",
  author    = "Santeramo, Ruggiero and Damiani, Celeste and Wei, Jiefei and
               Montana, Giovanni and Brentnall, Adam R",
  journal   = "Breast Cancer Res.",
  publisher = "Springer Science and Business Media LLC",
  volume    =  26,
  number    =  1,
  pages     = "25",
  month     =  feb,
  year      =  2024,
  copyright = "https://creativecommons.org/licenses/by/4.0",
  language  = "en"
}

@ARTICLE{Yala2021,
  title     = "Toward robust mammography-based models for breast cancer risk",
  author    = "Yala, Adam and Mikhael, Peter G and Strand, Fredrik and Lin,
               Gigin and Smith, Kevin and Wan, Yung-Liang and Lamb, Leslie and
               Hughes, Kevin and Lehman, Constance and Barzilay, Regina",
  journal   = "Sci. Transl. Med.",
  publisher = "American Association for the Advancement of Science (AAAS)",
  volume    =  13,
  number    =  578,
  pages     = "eaba4373",
  month     =  jan,
  year      =  2021,
  copyright = "https://www.sciencemag.org/about/science-licenses-journal-article-reuse",
  language  = "en"
}

@inproceedings{lomar,
  title={Longitudinal mammogram risk prediction},
  author={Karaman, Batuhan K and Dodelzon, Katerina and Akar, Gozde B and Sabuncu, Mert R},
  booktitle={International Conference on Medical Image Computing and Computer-Assisted Intervention},
  pages={437--446},
  year={2024},
  organization={Springer}
}

@inproceedings{vmra,
  title={VMRA-MaR: An Asymmetry-Aware Temporal Framework for Longitudinal Breast Cancer Risk Prediction},
  author={Sun, Zijun and Thrun, Solveig and Kampffmeyer, Michael},
  booktitle={International Conference on Medical Image Computing and Computer-Assisted Intervention},
  pages={660--669},
  year={2025},
  organization={Springer}
}

@misc{ccc,
  author = {Strand,  Fredrik},
  language = {en},
  title = {CSAW-CC (mammography) – a dataset for AI research to improve screening,  diagnostics and prognostics of breast cancer},
  publisher = {Karolinska Institutet},
  year = {2022}
}

@article{Dembrower2019,
  title = {A Multi-million Mammography Image Dataset and Population-Based Screening Cohort for the Training and Evaluation of Deep Neural Networks—the Cohort of Screen-Aged Women (CSAW)},
  volume = {33},
  ISSN = {1618-727X},
  number = {2},
  journal = {Journal of Digital Imaging},
  publisher = {Springer Science and Business Media LLC},
  author = {Dembrower,  Karin and Lindholm,  Peter and Strand,  Fredrik},
  year = {2019},
  month = sep,
  pages = {408–413}
}

@article{Reece2021,
  title = {Delayed or failure to follow-up abnormal breast cancer screening mammograms in primary care: a systematic review},
  volume = {21},
  ISSN = {1471-2407},
  number = {1},
  journal = {BMC Cancer},
  publisher = {Springer Science and Business Media LLC},
  author = {Reece,  Jeanette C. and Neal,  Eleanor F. G. and Nguyen,  Peter and McIntosh,  Jennifer G. and Emery,  Jon D.},
  year = {2021},
  month = apr 
}

@article{LaubySecretan2015,
  title = {Breast-Cancer Screening — Viewpoint of the IARC Working Group},
  volume = {372},
  ISSN = {1533-4406},
  number = {24},
  journal = {New England Journal of Medicine},
  publisher = {Massachusetts Medical Society},
  author = {Lauby-Secretan,  Béatrice and Scoccianti,  Chiara and Loomis,  Dana and Benbrahim-Tallaa,  Lamia and Bouvard,  Véronique and Bianchini,  Franca and Straif,  Kurt},
  year = {2015},
  month = jun,
  pages = {2353–2358}
}

@article{Smith2003,
  title = {Expert Group: IARC Handbooks of Cancer Prevention. Vol.7: Breast Cancer Screening. Lyon,  France: IARC; 2002. 248pp.: ISBN 92 832 3007 8},
  volume = {5},
  ISSN = {1465-542X},
  number = {4},
  journal = {Breast Cancer Research},
  publisher = {Springer Science and Business Media LLC},
  author = {Smith,  Robert A},
  year = {2003},
  month = aug 
}

@article{Ginsburg2020,
  title = {Breast cancer early detection: A phased approach to implementation},
  volume = {126},
  ISSN = {1097-0142},
  number = {S10},
  journal = {Cancer},
  publisher = {Wiley},
  author = {Ginsburg,  Ophira and Yip,  Cheng‐Har and Brooks,  Ari and Cabanes,  Anna and Caleffi,  Maira and Dunstan Yataco,  Jorge Antonio and Gyawali,  Bishal and McCormack,  Valerie and McLaughlin de Anderson,  Myrna and Mehrotra,  Ravi and Mohar,  Alejandro and Murillo,  Raul and Pace,  Lydia E. and Paskett,  Electra D. and Romanoff,  Anya and Rositch,  Anne F. and Scheel,  John R. and Schneidman,  Miriam and Unger‐Saldaña,  Karla and Vanderpuye,  Verna and Wu,  Tsu‐Yin and Yuma,  Safina and Dvaladze,  Allison and Duggan,  Catherine and Anderson,  Benjamin O.},
  year = {2020},
  month = apr,
  pages = {2379–2393}
}

@article{Grasso2025,
  title = {Innovative Methodologies for the Early Detection of Breast Cancer: A Review Categorized by Target Biological Samples},
  volume = {15},
  ISSN = {2079-6374},
  number = {4},
  journal = {Biosensors},
  publisher = {MDPI AG},
  author = {Grasso,  Antonella and Altomare,  Vittorio and Fiorini,  Giulia and Zompanti,  Alessandro and Pennazza,  Giorgio and Santonico,  Marco},
  year = {2025},
  month = apr,
  pages = {257}
}

@article{Karaman2022,
  title = {Machine learning based multi-modal prediction of future decline toward Alzheimer’s disease: An empirical study},
  volume = {17},
  ISSN = {1932-6203},
  number = {11},
  journal = {PLOS ONE},
  publisher = {Public Library of Science (PLoS)},
  author = {Karaman,  Batuhan K. and Mormino,  Elizabeth C. and Sabuncu,  Mert R.},
  editor = {Thung,  Kim Han},
  year = {2022},
  month = nov,
  pages = {e0277322}
}

@article{xin2022current,
  title={Do current multi-task optimization methods in deep learning even help?},
  author={Xin, Derrick and Ghorbani, Behrooz and Gilmer, Justin and Garg, Ankush and Firat, Orhan},
  journal={Advances in neural information processing systems},
  volume={35},
  pages={13597--13609},
  year={2022}
}

@article{vapnik2015,
author = {Vapnik, Vladimir and Izmailov, Rauf},
title = {Learning using privileged information: similarity control and knowledge transfer},
year = {2015},
issue_date = {January 2015},
publisher = {JMLR.org},
volume = {16},
number = {1},
issn = {1532-4435},
journal = {J. Mach. Learn. Res.},
month = jan,
pages = {2023–2049},
numpages = {27}
}

@article{vapnik2009,
author = {Vapnik, Vladimir and Vashist, Akshay},
title = {A new learning paradigm: Learning using privileged information},
year = {2009},
issue_date = {July, 2009},
publisher = {Elsevier Science Ltd.},
address = {GBR},
volume = {22},
number = {5–6},
issn = {0893-6080},
url = {https://doi.org/10.1016/j.neunet.2009.06.042},
doi = {10.1016/j.neunet.2009.06.042},
journal = {Neural Netw.},
month = jul,
pages = {544–557},
numpages = {14}
}

@article{himes2016breast,
  title={Breast cancer risk assessment: calculating lifetime risk using the Tyrer-Cuzick model},
  author={Himes, Deborah O and Root, Aubri E and Gammon, Amanda and Luthy, Karlen E},
  journal={The Journal for Nurse Practitioners},
  volume={12},
  number={9},
  pages={581--592},
  year={2016},
  publisher={Elsevier}
}

@article{tice2019validation,
  title={Validation of the breast cancer surveillance consortium model of breast cancer risk},
  author={Tice, Jeffrey A and Bissell, Michael CS and Miglioretti, Diana L and Gard, Charlotte C and Rauscher, Garth H and Dabbous, Firas M and Kerlikowske, Karla},
  journal={Breast cancer research and treatment},
  volume={175},
  number={2},
  pages={519--523},
  year={2019},
  publisher={Springer}
}

@misc{embed,
  doi = {10.48550/ARXIV.2202.04073},
  url = {https://arxiv.org/abs/2202.04073},
  author = {Jeong,  Jiwoong J. and Vey,  Brianna L. and Reddy,  Ananth and Kim,  Thomas and Santos,  Thiago and Correa,  Ramon and Dutt,  Raman and Mosunjac,  Marina and Oprea-Ilies,  Gabriela and Smith,  Geoffrey and Woo,  Minjae and McAdams,  Christopher R. and Newell,  Mary S. and Banerjee,  Imon and Gichoya,  Judy and Trivedi,  Hari},
  title = {The EMory BrEast imaging Dataset (EMBED): A Racially Diverse,  Granular Dataset of 3.5M Screening and Diagnostic Mammograms},
  publisher = {arXiv},
  year = {2022},
  copyright = {Creative Commons Attribution 4.0 International}
}

@article{omidb,
  title = {OPTIMAM Mammography Image Database: A Large-Scale Resource of                    Mammography Images and Clinical Data},
  volume = {3},
  ISSN = {2638-6100},
  url = {http://dx.doi.org/10.1148/ryai.2020200103},
  DOI = {10.1148/ryai.2020200103},
  number = {1},
  journal = {Radiology: Artificial Intelligence},
  publisher = {Radiological Society of North America (RSNA)},
  author = {Halling-Brown,  Mark D. and Warren,  Lucy M. and Ward,  Dominic and Lewis,  Emma and Mackenzie,  Alistair and Wallis,  Matthew G. and Wilkinson,  Louise S. and Given-Wilson,  Rosalind M. and McAvinchey,  Rita and Young,  Kenneth C.},
  year = {2021},
  month = Jan,
  pages = {e200103}
}

@article{lmv,
  title={Longitudinal Multi-View Breast Cancer Risk Prediction},
  author={Thrun, Solveig and Sun, Zijun and Salahuddin, Suaiba A and Wickstr{\o}m, Kristoffer and Wetzer, Elisabeth and Hansen, Stine and Jenssen, Robert and Kampffmeyer, Michael},
  journal={arXiv preprint arXiv:2607.11343},
  year={2026}
}

@article{vmamba24,
  title={Vision mamba: Efficient visual representation learning with bidirectional state space model},
  author={Zhu, Lianghui and Liao, Bencheng and Zhang, Qian and Wang, Xinlong and Liu, Wenyu and Wang, Xinggang},
  journal={arXiv preprint arXiv:2401.09417},
  year={2024}
}

@article{wang2025mtp,
  title={Predicting short-to long-term breast cancer risk from longitudinal mammographic screening history},
  author={Wang, Xin and Tan, Tao and Gao, Yuan and Su, Ruisheng and Teuwen, Jonas and Kroes, Jaap and Zhang, Tianyu and D’Angelo, Anna and Han, Luyi and Drukker, Caroline A and others},
  journal={NPJ Breast Cancer},
  volume={11},
  number={1},
  pages={118},
  year={2025},
  publisher={Nature Publishing Group UK London}
}

@INPROCEEDINGS{mha,
  author={Aslam, Muhammad Haseeb and Osama Zeeshan, Muhammad and Pedersoli, Marco and Koerich, Alessandro L. and Bacon, Simon and Granger, Eric},
  booktitle={2023 IEEE/CVF Conference on Computer Vision and Pattern Recognition Workshops (CVPRW)}, 
  title={Privileged Knowledge Distillation for Dimensional Emotion Recognition in the Wild}, 
  year={2023},
  volume={},
  number={},
  pages={3338-3347},
  doi={10.1109/CVPRW59228.2023.00336}}

@inproceedings{aslam2024distilling,
  title={Distilling privileged multimodal information for expression recognition using optimal transport},
  author={Aslam, Muhammad Haseeb and Zeeshan, Muhammad Osama and Belharbi, Soufiane and Pedersoli, Marco and Koerich, Alessandro Lameiras and Bacon, Simon and Granger, Eric},
  booktitle={2024 IEEE 18th International Conference on Automatic Face and Gesture Recognition (FG)},
  pages={1--10},
  year={2024},
  organization={IEEE}
}

@article{aslam2024multi,
  title={Multi teacher privileged knowledge distillation for multimodal expression recognition},
  author={Aslam, Muhammad Haseeb and Pedersoli, Marco and Koerich, Alessandro Lameiras and Granger, Eric},
  journal={arXiv preprint arXiv:2408.09035},
  year={2024}
}

@article{
provodin2025rethinking,
title={Rethinking Knowledge Transfer in Learning Using Privileged Information},
author={Danil Provodin and Bram van den Akker and Christina Katsimerou and Maurits Clemens Kaptein and Mykola Pechenizkiy},
journal={Transactions on Machine Learning Research},
issn={2835-8856},
year={2025},
url={https://openreview.net/forum?id=dg1tqNIWg3},
note={}
}

@inproceedings{lopezpaz2016unifying,
  author       = {David Lopez{-}Paz and
                  L{\'{e}}on Bottou and
                  Bernhard Sch{\"{o}}lkopf and
                  Vladimir Vapnik},
  editor       = {Yoshua Bengio and
                  Yann LeCun},
  title        = {Unifying distillation and privileged information},
  booktitle    = {4th International Conference on Learning Representations, {ICLR} 2016,
                  San Juan, Puerto Rico, May 2-4, 2016, Conference Track Proceedings},
  year         = {2016},
  url          = {http://arxiv.org/abs/1511.03643},
  bibsource    = {dblp computer science bibliography, https://dblp.org}
}

@article{
hong2025mutud,
title={Efficient Audiovisual Speech Processing via {MUTUD}: Multimodal Training and Unimodal Deployment},
author={Joanna Hong and Sanjeel Parekh and Honglie Chen and Jacob Donley and Ke Tan and Buye Xu and Anurag Kumar},
journal={Transactions on Machine Learning Research},
issn={2835-8856},
year={2026},
url={https://openreview.net/forum?id=5bshBY8RDf},
note={}
}

@article{wang2020,
  title={Multi-frame to single-frame: Knowledge distillation for 3d object detection},
  author={Wang, Yue and Fathi, Alireza and Wu, Jiajun and Funkhouser, Thomas and Solomon, Justin},
  journal={arXiv preprint arXiv:2009.11859},
  year={2020}
}

@INPROCEEDINGS{qiu2023,
  author={Qiu, Shoumeng and Jiang, Feng and Zhang, Haiqiang and Xue, Xiangyang and Pu, Jian},
  booktitle={2023 IEEE International Conference on Robotics and Automation (ICRA)}, 
  title={Multi-to-Single Knowledge Distillation for Point Cloud Semantic Segmentation}, 
  year={2023},
  volume={},
  number={},
  pages={9303-9309},
  doi={10.1109/ICRA48891.2023.10160496}}

@article{davydov2024using,
  title={Using Motion Cues to Supervise Single-frame Body Pose \& Shape Estimation in Low Data Regimes},
  author={Andrey Davydov and Alexey Sidnev and Artsiom Sanakoyeu and Yuhua Chen and Mathieu Salzmann and Pascal Fua},
  journal={Trans. Mach. Learn. Res.},
  year={2024},
  volume={2024},
  url={https://api.semanticscholar.org/CorpusID:271770092}
}

@InProceedings{karlsson2021using,
  title = 	 { Using time-series privileged information for provably efficient learning of prediction models },
  author =       {K.A. Karlsson, Rickard and Willbo, Martin and Hussain, Zeshan M. and Krishnan, Rahul G. and Sontag, David and Johansson, Fredrik},
  booktitle = 	 {Proceedings of The 25th International Conference on Artificial Intelligence and Statistics},
  pages = 	 {5459--5484},
  year = 	 {2022},
  editor = 	 {Camps-Valls, Gustau and Ruiz, Francisco J. R. and Valera, Isabel},
  volume = 	 {151},
  series = 	 {Proceedings of Machine Learning Research},
  month = 	 {28--30 Mar},
  publisher =    {PMLR},
  url = 	 {https://proceedings.mlr.press/v151/k-a-karlsson22a.html}
}
\bibliographystyle{tmlr}
\clearpage
\appendix
\appendix
\section{Additional Experimental Details and Analyses}
\label{app:additional-analyses}


\subsection{Detailed Cohort Characteristics}
\label{app:cohort-details}
Table~\ref{tab:dataset-statistics-detailed} summarizes the final experimental cohorts after eligibility filtering and examination-level duplication. The datasets differ substantially in cohort size, available history length, temporal coverage, and future cancer prevalence, providing complementary settings for evaluating \ours.

CSAW-CC contains 8,723 patients and 24,694 examinations, with 75.50\% of patients contributing at least two examinations. Four examinations per patient is the most common history length, and the cohort provides temporal spans of up to eight calendar years. EMBED is the largest cohort, containing 16,487 patients and 42,369 examinations, although 43.86\% of its patients have only one examination. Its longitudinal subset includes as many as nine examinations per patient and spans up to 7.59 years.

The OMI-DB cohort subset contains 11,531 patients and 36,859 examinations, with all retained patients contributing at least two examinations. Most OMI-DB patients have three examinations, while the maximum first-to-last span reaches 13 years. Future malignant prevalence also differs considerably across datasets, ranging from 2.44\% in EMBED to 22.40\% in OMI-DB. These differences demonstrate that the three cohorts provide heterogeneous longitudinal settings rather than variations of a single data distribution.

\begin{table}[H]
\centering
\caption{Detailed longitudinal cohort statistics. Counts in parentheses denote the percentage of patients in the corresponding dataset. For CSAW-CC, temporal span is measured as the difference between calendar examination years. For EMBED and OMI-DB, elapsed years are computed from exact examination dates and grouped using the floor of the elapsed number of years.}
\label{tab:dataset-statistics-detailed}
\scriptsize
\setlength{\tabcolsep}{5pt}

\textbf{(a) Cohort-level statistics}
\vspace{2pt}

\resizebox{\textwidth}{!}{
\begin{tabular}{lrrrrrrr}
\toprule
Dataset & \# Patients & \# Exams & One exam & At least two exams & Max. exams & Max. first--last span & \# Future malignant patients \\
\midrule
CSAW-CC & 8,723 & 24,694 & 2,137 (24.50\%) & 6,586 (75.50\%) & 6 & 8 years & 533 (6.11\%) \\
EMBED & 16,487 & 42,369 & 7,232 (43.86\%) & 9,255 (56.14\%) & 9 & 2,774 days (7.59 years) & 402 (2.44\%) \\
OMI-DB & 11,531 & 36,859 & 0 (0.00\%) & 11,531 (100.00\%) & 9 & 5,016 days (13.73 years) & 2,583 (22.40\%) \\
\bottomrule
\end{tabular}
}

\vspace{0.8em}

\begin{minipage}[t]{0.49\textwidth}
\centering
\textbf{(b) Number of examinations per patient}
\vspace{2pt}

\setlength{\tabcolsep}{3.5pt}
\begin{tabular}{rccc}
\toprule
\# Exams & CSAW-CC & EMBED & OMI-DB \\
\midrule
1 & 2,137 (24.50\%) & 7,232 (43.86\%) & -- \\
2 & 1,250 (14.33\%) & 3,089 (18.74\%) & 1,859 (16.12\%) \\
3 & 1,770 (20.29\%) & 1,878 (11.39\%) & 6,200 (53.77\%) \\
4 & 3,086 (35.38\%) & 1,316 (7.98\%) & 2,859 (24.79\%) \\
5 & 477 (5.47\%) & 941 (5.71\%) & 581 (5.04\%) \\
6 & 3 (0.03\%) & 996 (6.04\%) & 28 (0.24\%) \\
7 & -- & 902 (5.47\%) & 1 (0.01\%) \\
8 & -- & 131 (0.79\%) & 2 (0.02\%) \\
9 & -- & 2 (0.01\%) & 1 (0.01\%) \\
\bottomrule
\end{tabular}
\end{minipage}
\hfill
\begin{minipage}[t]{0.49\textwidth}
\centering
\textbf{(c) First-to-last examination span}
\vspace{2pt}

\setlength{\tabcolsep}{3.5pt}
\begin{tabular}{rccc}
\toprule
Span (years) & CSAW-CC & EMBED & OMI-DB \\
\midrule
0  & 2,137 (24.50\%) & 7,491 (45.44\%) & 15 (0.13\%) \\
1  & 162 (1.86\%) & 1,957 (11.87\%) & 86 (0.75\%) \\
2  & 655 (7.51\%) & 1,386 (8.41\%) & 1,434 (12.44\%) \\
3  & 503 (5.77\%) & 1,239 (7.52\%) & 464 (4.02\%) \\
4  & 1,133 (12.99\%) & 1,165 (7.07\%) & 325 (2.82\%) \\
5  & 1,319 (15.12\%) & 1,464 (8.88\%) & 4,303 (37.32\%) \\
6  & 1,631 (18.70\%) & 1,540 (9.34\%) & 1,344 (11.66\%) \\
7  & 1,087 (12.46\%) & 245 (1.49\%) & 226 (1.96\%) \\
8  & 96 (1.10\%) & -- & 1,699 (14.73\%) \\
9  & -- & -- & 958 (8.31\%) \\
10 & -- & -- & 83 (0.72\%) \\
11 & -- & -- & 556 (4.82\%) \\
12 & -- & -- & 34 (0.29\%) \\
13 & -- & -- & 4 (0.03\%) \\
\bottomrule
\end{tabular}
\end{minipage}

\vspace{4pt}
\parbox{\textwidth}{\footnotesize A dash indicates that no patients occurred in the corresponding category. A zero-year span may include multiple examinations separated by less than one year for EMBED and OMI-DB because elapsed durations are grouped using floor binning. We only pick one of those. Future malignant counts follow the cohort-specific outcome construction described in the main text. In CSAW-CC, 873 patients had at least one cancer examination, of whom 533 had a cancer examination after their first retained examination. For OMI-DB, the corresponding counts were 2,781 and 2,583 patients.}
\end{table}


\subsection{Detailed Missing-History Results}
\label{app:missing-history-results}
To determine whether the gains from \ours arise merely from populating missing temporal positions, we compare it with three alternative strategies using the VMRA backbone. \textit{Masked Training} randomly removes observed historical visits during training to improve robustness to missing history. \textit{All $\VisitEmbedding{0}$} replaces every missing prior-visit embedding with the current-visit embedding, whereas \textit{All Mean} replaces each missing prior-visit embedding with the corresponding year-specific mean embedding computed from the available training histories. Thus, a separate mean is computed for each prior-visit position rather than averaging embeddings across all historical years. The standard VMRA model evaluated with all prior visits removed ($\eta=0$) is also included as the single-exam baseline. All methods operate under the same current-exam-only inference constraint and have no access to true prior mammograms at deployment.

Across the shorter 1--3 year horizons, all missing-history strategies perform similarly, with only minor differences relative to the single-exam VMRA baseline ($\eta=0$). Masked training and simple history substitutions such as All $\VisitEmbedding{0}$ therefore provide little advantage over single-exam inference at these horizons. In contrast, the benefit of \ours becomes pronounced at longer horizons: at 4 and 5 years, it achieves the highest AUC and pAUC, with the largest improvement observed at 5 years. This pattern suggests that the transferred longitudinal information is particularly beneficial for longer-term risk prediction, where removing true screening history has a larger impact on the underlying longitudinal model.

\begin{table*}[t]
\centering
\small
\caption{Comparison with alternative missing-history strategies for VMRA on CSAW-CC across all prediction horizons. All methods are evaluated without access to true prior mammograms at inference. Results are reported as mean$\pm$std across 10 patient-level splits.}
\label{tab:missing_baselines}
\setlength{\tabcolsep}{8pt}

\begin{tabular}{clccccc}
\toprule
& Method & 1-Year & 2-Year & 3-Year & 4-Year & 5-Year \\
\midrule

\multirow{5}{*}{\rotatebox[origin=c]{90}{AUC}}
& VMRA ($\eta=0$)
& \textbf{0.922$\pm$0.020}
& 0.872$\pm$0.017
& \textbf{0.853$\pm$0.016}
& 0.836$\pm$0.015
& 0.829$\pm$0.015 \\

& Masked Training
& 0.922$\pm$0.021
& \textbf{0.873$\pm$0.017}
& \textbf{0.853$\pm$0.016}
& 0.837$\pm$0.015
& 0.829$\pm$0.016 \\

& All $\VisitEmbedding{0}$
& 0.922$\pm$0.021
& 0.872$\pm$0.016
& 0.852$\pm$0.016
& 0.832$\pm$0.016
& 0.819$\pm$0.021 \\

& All Mean
& 0.912$\pm$0.023
& 0.860$\pm$0.019
& 0.834$\pm$0.020
& 0.814$\pm$0.021
& 0.814$\pm$0.022 \\

& \ours (Ours)
& 0.920$\pm$0.018
& 0.869$\pm$0.018
& 0.852$\pm$0.016
& \textbf{0.847$\pm$0.015}
& \textbf{0.855$\pm$0.017} \\

\midrule

\multirow{5}{*}{\rotatebox[origin=c]{90}{pAUC@10\%}}
& VMRA ($\eta=0$)
& 0.823$\pm$0.023
& 0.752$\pm$0.019
& 0.734$\pm$0.018
& 0.714$\pm$0.019
& 0.710$\pm$0.021 \\

& Masked Training
& 0.823$\pm$0.024
& 0.754$\pm$0.021
& \textbf{0.735$\pm$0.019}
& 0.715$\pm$0.020
& 0.711$\pm$0.022 \\

& All $\VisitEmbedding{0}$
& \textbf{0.824$\pm$0.025}
& \textbf{0.754$\pm$0.020}
& 0.734$\pm$0.020
& 0.711$\pm$0.021
& 0.689$\pm$0.040 \\

& All Mean
& 0.816$\pm$0.028
& 0.746$\pm$0.023
& 0.719$\pm$0.021
& 0.692$\pm$0.022
& 0.669$\pm$0.047 \\

& \ours (Ours)
& 0.818$\pm$0.020
& 0.749$\pm$0.017
& 0.733$\pm$0.017
& \textbf{0.734$\pm$0.017}
& \textbf{0.757$\pm$0.018} \\

\bottomrule
\end{tabular}
\end{table*}
Masked Training provides only negligible gains over the standard no-history VMRA baseline, increasing 5-year pAUC@10\% from 0.710 to 0.711 while leaving AUC unchanged. Heuristic embedding replacement performs worse: using the current embedding for every missing visit reduces 5-year AUC and pAUC@10\% to 0.819 and 0.689, while mean-embedding replacement further reduces them to 0.814 and 0.669.

\ours achieves the strongest result for both metrics and horizons. Relative to the no-history VMRA baseline, it increases 4-year AUC from 0.836 to 0.847 and pAUC@10\% from 0.714 to 0.734. At five years, AUC improves from 0.829 to 0.855 and pAUC@10\% from 0.710 to 0.757. These findings show that \ours does more than populate missing temporal positions: patient-specific latent history prediction and privileged teacher supervision recover information that is not captured by masking or fixed embedding replacement.


\subsection{Optimization of the Historical Embedding Predictor}
\label{app:predictor-optimization}

\begin{figure*}[t]
\centering
\includegraphics[width=0.7\textwidth]{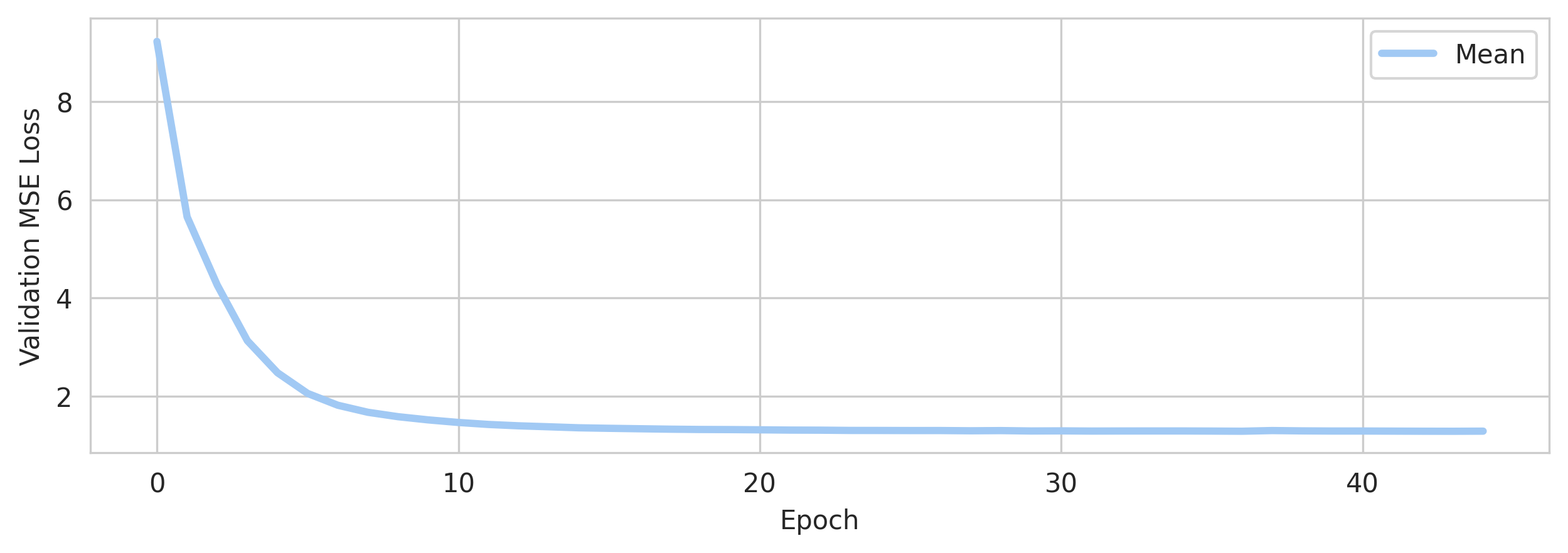}
\caption{Validation MSE of the historical embedding predictor. The loss decreases steadily and stabilizes after approximately 15 epochs, indicating stable optimization of predicted historical visit embeddings.}
\label{fig:val_mse}
\end{figure*}

Figure~\ref{fig:val_mse} reports the validation mean squared error of the historical embedding predictor. The loss decreases steadily during training and stabilizes after approximately 15 epochs, indicating stable optimization of the feature-prediction objective.

The predictor is not trained to reconstruct historical mammograms or reproduce their visual details. Instead, it estimates compact visit-level representations in the frozen encoder feature space. The convergence of the validation loss therefore indicates that the current examination contains sufficient signal to approximate the target historical embeddings under the adopted feature-level objective. Convergence alone does not establish the clinical utility of these representations; their usefulness is instead supported by the downstream improvements in long-horizon AUC and pAUC reported in the main experiments.


\subsection{Computational cost.}
\label{app:computational-cost}
Table~\ref{tab:computational-cost} compares the computational requirements of LoMaR and LoMaR+\ours. At deployment, \ours increases the parameter count from 1.474M to 3.305M and the forward-pass complexity from 0.0074 to 0.0092 GFLOPs per sample. Measured inference latency increases from $1.960$ to $4.182$ ms per batch. Training incurs additional overhead from the horizon-specific teachers, increasing the measured training-step time from $13.065$ to $26.255$ ms per batch; the complete training model contains 7.705M parameters including the teacher networks. Importantly, the teachers are used only for distillation during training and are discarded at deployment, so their parameters and computation do not contribute to inference.
\begin{table}[t]
\centering
\caption{Computational cost of LoMaR and LoMaR+\ours. Training latency includes forward and backward propagation and the optimizer update; for \ours, it also includes teacher-based distillation. Inference measurements and GFLOPs correspond to the deployed model, for which teacher networks are removed. Runtime is measured on an otherwise idle NVIDIA RTX A6000 using CUDA events. The batch size is 1 for this.}
\label{tab:computational-cost}
\begin{tabular}{lrrrr}
\toprule
Method & Params (M) & GFLOPs & Train (ms/batch) & Infer. (ms/batch) \\
\midrule
LoMaR          & 1.474 & 0.0074 & $13.065 \pm 3.981$ & $1.960 \pm 0.010$ \\
LoMaR+\ours    & 3.305 & 0.0092 & $26.255 \pm 1.661$ & $4.182 \pm 0.029$ \\
\bottomrule
\end{tabular}
\end{table}

\subsection{Complete AUC Significance Comparisons}
\label{app:statistical-comparisons}

Table~\ref{tab:auc-significance-full} provides the complete statistical comparisons underlying the significance analysis in the main text. For each dataset and \ours backbone, we compare \ours with Mirai, the corresponding model supplied with observed screening history at inference ($\eta=4$), and the corresponding longitudinally trained model evaluated without history ($\eta=0$). Statistical significance is assessed using $t$-test p-values across the 10 patient-level splits, with values below $0.05$ shown in bold. Relative to the corresponding missing-history baselines, \ours significantly improves 5-year AUC for both backbones on all three datasets. On OMI-DB, all comparisons with the corresponding missing-history baselines are significant across the five prediction horizons, while at the 4-year horizon all missing-history comparisons are significant except LoMaR+\ours on CSAW-CC. Comparisons with Mirai show a broadly similar dataset- and horizon-dependent pattern. Relative to the full-history references, LoMaR+\ours shows no significant difference on CSAW-CC, whereas VMRA+\ours differs significantly at the 4- and 5-year horizons. On OMI-DB, both \ours variants differ significantly from their corresponding full-history references across all five horizons. On EMBED, LoMaR+\ours differs significantly from full-history LoMaR at 2--4 years, while no significant differences are observed between VMRA+\ours and full-history VMRA.
\begin{table*}[t]
\centering
\scriptsize
\setlength{\tabcolsep}{3.8pt}
\renewcommand{\arraystretch}{1.08}

\caption{
$t$-test $p$-values for AUC across 10 patient-level splits. Each row compares the indicated reference model with the corresponding \ours model. $\eta=4$ denotes inference with observed screening history, whereas $\eta=0$ denotes missing-history inference. Bold values indicate $p<0.05$.
}
\label{tab:auc-significance-full}

\resizebox{0.98\textwidth}{!}{
\begin{tabular}{c l l c c c c c}
\toprule
Dataset & \ours model & Reference model & 1y & 2y & 3y & 4y & 5y \\
\midrule

\multirow{6}{*}{CSAW-CC}
& \multirow{3}{*}{LoMaR+\ours}
& Mirai
& \textbf{0.001789}
& \textbf{0.037618}
& 0.217512
& 0.171142
& \textbf{2.820e-07} \\

&
& LoMaR ($\eta=4$)
& 0.152219
& 0.179205
& 0.795078
& 0.239887
& 0.266281 \\

&
& LoMaR ($\eta=0$)
& \textbf{0.004475}
& \textbf{0.017474}
& 0.280574
& 0.133718
& \textbf{1.028e-07} \\

\cmidrule(lr){2-8}

&
\multirow{3}{*}{VMRA+\ours}
& Mirai
& \textbf{0.004399}
& 0.204405
& \textbf{0.041503}
& \textbf{5.941e-06}
& \textbf{5.116e-09} \\

&
& VMRA ($\eta=4$)
& 0.358247
& 0.348782
& 0.112775
& \textbf{0.000102}
& \textbf{0.000787} \\

&
& VMRA ($\eta=0$)
& \textbf{0.011003}
& \textbf{0.034079}
& 0.158398
& \textbf{6.288e-06}
& \textbf{6.293e-09} \\

\midrule

\multirow{6}{*}{OMI-DB}
& \multirow{3}{*}{LoMaR+\ours}
& Mirai
& \textbf{2.424e-11}
& \textbf{3.062e-11}
& \textbf{1.346e-07}
& \textbf{3.764e-11}
& \textbf{4.336e-11} \\

&
& LoMaR ($\eta=4$)
& \textbf{5.860e-05}
& \textbf{6.033e-05}
& \textbf{0.000339}
& \textbf{0.000133}
& \textbf{0.000134} \\

&
& LoMaR ($\eta=0$)
& \textbf{1.448e-10}
& \textbf{1.428e-10}
& \textbf{5.937e-07}
& \textbf{5.156e-10}
& \textbf{3.813e-10} \\

\cmidrule(lr){2-8}

&
\multirow{3}{*}{VMRA+\ours}
& Mirai
& \textbf{9.378e-11}
& \textbf{1.801e-10}
& \textbf{3.509e-07}
& \textbf{4.077e-10}
& \textbf{2.443e-10} \\

&
& VMRA ($\eta=4$)
& \textbf{0.000961}
& \textbf{0.001282}
& \textbf{0.012227}
& \textbf{0.016342}
& \textbf{0.016968} \\

&
& VMRA ($\eta=0$)
& \textbf{1.203e-10}
& \textbf{3.013e-10}
& \textbf{9.315e-07}
& \textbf{5.224e-10}
& \textbf{2.447e-10} \\

\midrule

\multirow{6}{*}{EMBED}
& \multirow{3}{*}{LoMaR+\ours}
& Mirai
& 0.050205
& 0.631398
& \textbf{0.021478}
& \textbf{0.008170}
& \textbf{0.006850} \\

&
& LoMaR ($\eta=4$)
& 0.062239
& \textbf{0.014575}
& \textbf{0.021120}
& \textbf{0.003020}
& 0.393492 \\

&
& LoMaR ($\eta=0$)
& \textbf{0.008841}
& 0.200104
& 0.051954
& \textbf{0.003385}
& \textbf{0.006718} \\

\cmidrule(lr){2-8}

&
\multirow{3}{*}{VMRA+\ours}
& Mirai
& 0.463843
& 0.379743
& 0.823028
& \textbf{0.013211}
& \textbf{0.001438} \\

&
& VMRA ($\eta=4$)
& 0.458128
& 0.873566
& 0.409576
& 0.067134
& 0.320490 \\

&
& VMRA ($\eta=0$)
& \textbf{0.044933}
& \textbf{0.049366}
& 0.711080
& \textbf{0.011380}
& \textbf{0.000643} \\

\bottomrule
\end{tabular}
}
\end{table*}
\subsection{Single-Teacher versus Multi-Teacher Distillation}
\label{app:single-vs-multi}

\begin{figure}[t]
    \centering
    \includegraphics[width=\linewidth]{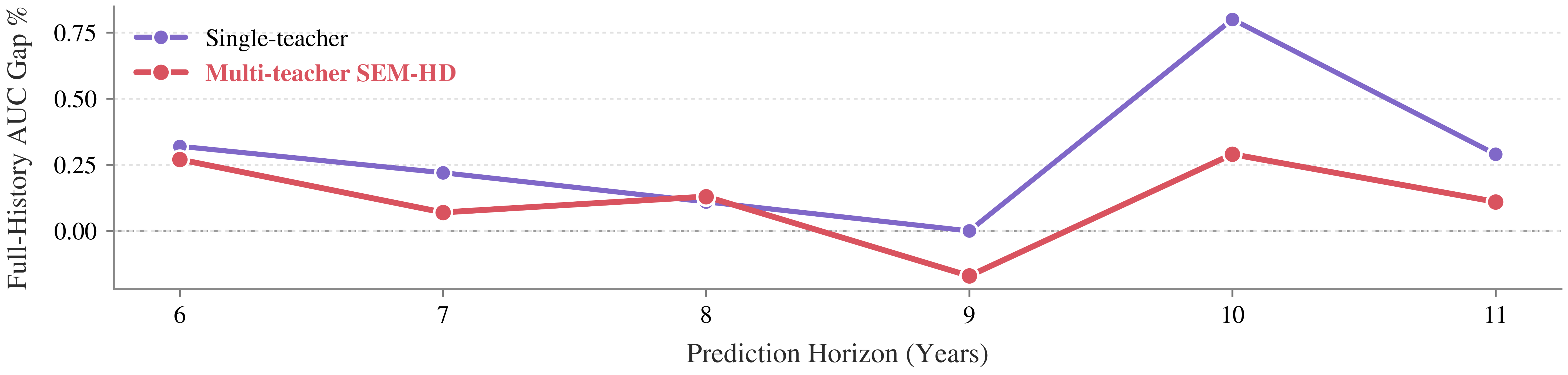}
    \caption{Comparison of single-teacher and horizon-specific multi-teacher distillation on OMI-DB using the LoMaR backbone. The plot shows the AUC gap relative to full-history inference for prediction horizons from 6 to 11 years. Lower values indicate closer agreement with the full-history reference, and negative values indicate that the student numerically exceeds it. The multi-teacher \ours variant achieves an equal or smaller gap at most evaluated horizons, while the single-teacher variant remains a competitive lower-cost alternative.}
    \label{fig:long-horizon-teacher-ablation}
\end{figure}

To examine the tradeoff between teacher-training cost and predictive performance, we compare the proposed horizon-specific multi-teacher formulation with a reduced-cost variant using a single teacher. Figure~\ref{fig:long-horizon-teacher-ablation} reports the AUC gap to full-history inference from 6 to 11 years on OMI-DB, where smaller values indicate closer agreement with the full-history reference.

Both variants remain close to full-history performance across the extended horizons, showing that even a single teacher can transfer substantial longitudinal information to the current-exam-only student. However, the proposed multi-teacher formulation achieves an equal or smaller gap at five of the six evaluated horizons, with the clearest advantage at 10 years. This suggests that horizon-specific teachers provide complementary supervision that improves consistency across prediction horizons, while the single-teacher variant offers a lower-cost alternative when teacher-training resources are constrained. Importantly, all teacher models are used only during training and are discarded at inference, so this tradeoff affects training cost but not deployment-time complexity.

\subsection{Hyperparameter sensitivity analysis}
\label{app:hp}

\begin{figure*}[t]
    \centering
    \includegraphics[width=\textwidth]{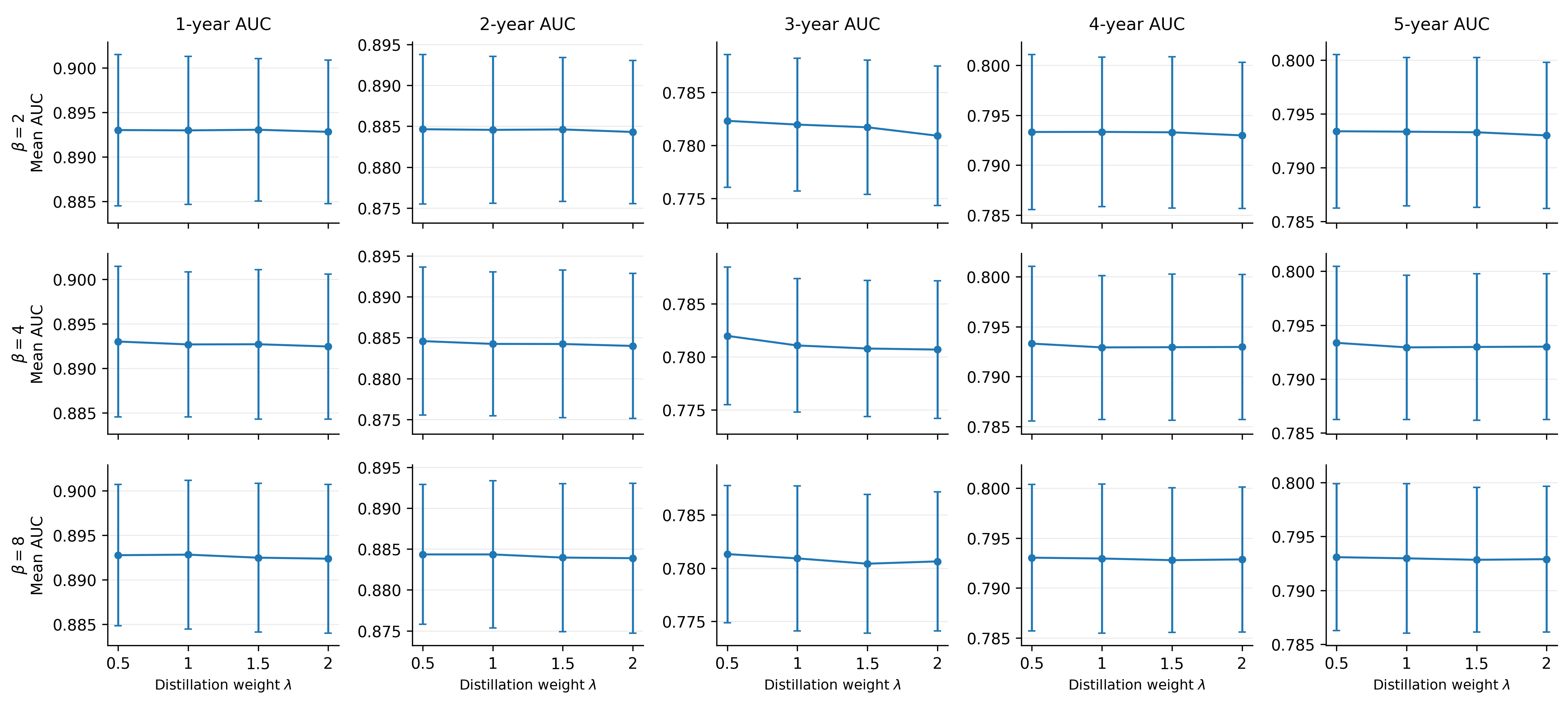}
    \caption{Hyperparameter sensitivity of LoMaR+\ours on OMI-DB. Rows correspond to the distillation temperature $\beta\in\{2,4,8\}$ and columns correspond to the 1--5 year prediction horizons. The x-axis shows the distillation weight $\lambda$, while each point and error bar denote the mean and standard deviation of AUC across 10 splits. The results remain stable across the tested hyperparameter range, indicating that \ours is robust to moderate changes in $\lambda$ and $\beta$.}
    \label{fig:hyper_sensitivity}
\end{figure*}

To assess the robustness of \ours to its main distillation hyperparameters, Fig.~\ref{fig:hyper_sensitivity} reports the performance of LoMaR+\ours on OMI-DB under different values of the distillation weight $\lambda$ and temperature $\beta$. The rows correspond to $\beta\in\{2,4,8\}$, the columns correspond to the 1-5 year prediction horizons, and the x-axis in each panel shows $\lambda\in\{0.5,1.0,1.5,2.0\}$. Each point denotes the mean AUC over 10 splits and the error bars indicate the standard deviation.

Overall, the results show that \ours is robust to moderate changes in either hyperparameter. Across all horizons, the mean AUC varies only slightly as $\lambda$ and $\beta$ change, and these differences are small relative to the split-to-split variation. This trend is particularly clear at the clinically important 4- and 5-year horizons, where the performance remains stable across all tested settings. These results suggest that the proposed method does not rely on narrow hyperparameter tuning and behaves robustly over a reasonable range of distillation configurations.

\end{document}